\pdfoutput=1
\documentclass{article} %
\PassOptionsToPackage{table}{xcolor}
\usepackage{iclr2027_conference,times}

\usepackage{amsmath,amsfonts,bm}

\def\eqref#1{equation~\ref{#1}}

\def\1{\bm{1}}

\DeclareMathAlphabet{\mathsfit}{\encodingdefault}{\sfdefault}{m}{sl}
\SetMathAlphabet{\mathsfit}{bold}{\encodingdefault}{\sfdefault}{bx}{n}

\usepackage{hyperref}
\usepackage{url}

\usepackage{latexsym}
\usepackage[T1]{fontenc}
\usepackage[utf8]{inputenc}
\usepackage{microtype}
\usepackage{xcolor}
\usepackage{graphicx}
\usepackage{booktabs}
\usepackage{amsfonts}
\usepackage{nicefrac}
\usepackage{enumitem}

\usepackage{tikz}
\usetikzlibrary{arrows.meta, positioning}
\usepackage{amsmath}
\usepackage{makecell} %

\usepackage{tabularx}
\usepackage{afterpage}
\usepackage{wrapfig}
\usepackage{siunitx}
\usepackage{threeparttable}
\usepackage{comment}
\usepackage{multirow}

\IfFileExists{algorithm.sty}{
  \usepackage{algorithm}
}{
  \usepackage{float}
  \floatstyle{ruled}
  \newfloat{algorithm}{tbp}{loa}
  \floatname{algorithm}{Algorithm}
}

\makeatletter
\newcommand{\apxcaptionof}[1]{\def\@captype{#1}\caption}
\makeatother

\title{Towards Mechanistically Understanding Why Memorized Knowledge Fails to Generalize in Large Language Model Finetuning}

\author{%
  \textbf{Lu Dai\textsuperscript{1,2}} \quad
  \textbf{Ziyang Rao\textsuperscript{2}} \quad
  \textbf{Yili Wang\textsuperscript{2}} \quad
  \textbf{Hanqing Wang\textsuperscript{2}} \\
  \textbf{Hao Liu\textsuperscript{2,1}} \quad
  \textbf{Hui Xiong\textsuperscript{2,1}} \\
  \textsuperscript{1}HKUST \quad
  \textsuperscript{2}HKUST(GZ) \\
  \texttt{ldaiae@connect.ust.hk} \quad
  \texttt{\{liuh,xionghui\}@ust.hk}
}

\iclrfinalcopy
\begin{document}
\maketitle
\lhead{Preprint}

\begin{abstract}
Fine-tuning LLMs to inject new knowledge faces a critical challenge: LLMs can quickly memorize new facts, yet fail to use them for downstream reasoning tasks.
We formalize this failure as the \textit{\textbf{Knowing--Using Gap}}, characterized by an accuracy gap and a temporal lag between memorization and generalization.
To understand this phenomenon, we fine-tune LLMs with unseen knowledge and monitor the spatial permeation dynamics of the knowledge internally with self-patching, an adaptation of activation patching that scans all layer pairs at every fine-tuning checkpoint.
Self-patching identifies activation locations where relocating representations substantially improves failed generalization cases. These results are consistent with a knowledge-circuit misalignment hypothesis: memorized representations can exist internally but may not be routed to computation-effective layers.
Experiments are done cross-domain for the robustness of this finding.
Building on this diagnosis, we propose layer-wise representation self-distillation (LRSD) that aligns knowledge representation from late storage layer to middle layer. 
LRSD keeps improving generalization after fine-tuning saturates and nearly doubles multi-hop chaining accuracy on Qwen, while leaving memorization intact.
Code and data are open at \url{https://anonymous.4open.science/r/Mem2Gen-71FF}.
\end{abstract}

\section{Introduction}

Large language models (LLMs) excel at varieties of tasks but face significant challenges in adapting to unseen information, necessitating effective methods for post-training knowledge updates. While there are approaches like retrieval-augmented generation (RAG) and knowledge editing \cite{memit2023, gupta2024model}, fine-tuning remains a fundamental paradigm for knowledge updating, as it not only operates on parametric memory end-to-end but also injects knowledge in a way that in principle allows reuse by the model’s existing reasoning capabilities.

\begin{figure*}[ht]
    \centering
    \includegraphics[width=\columnwidth]{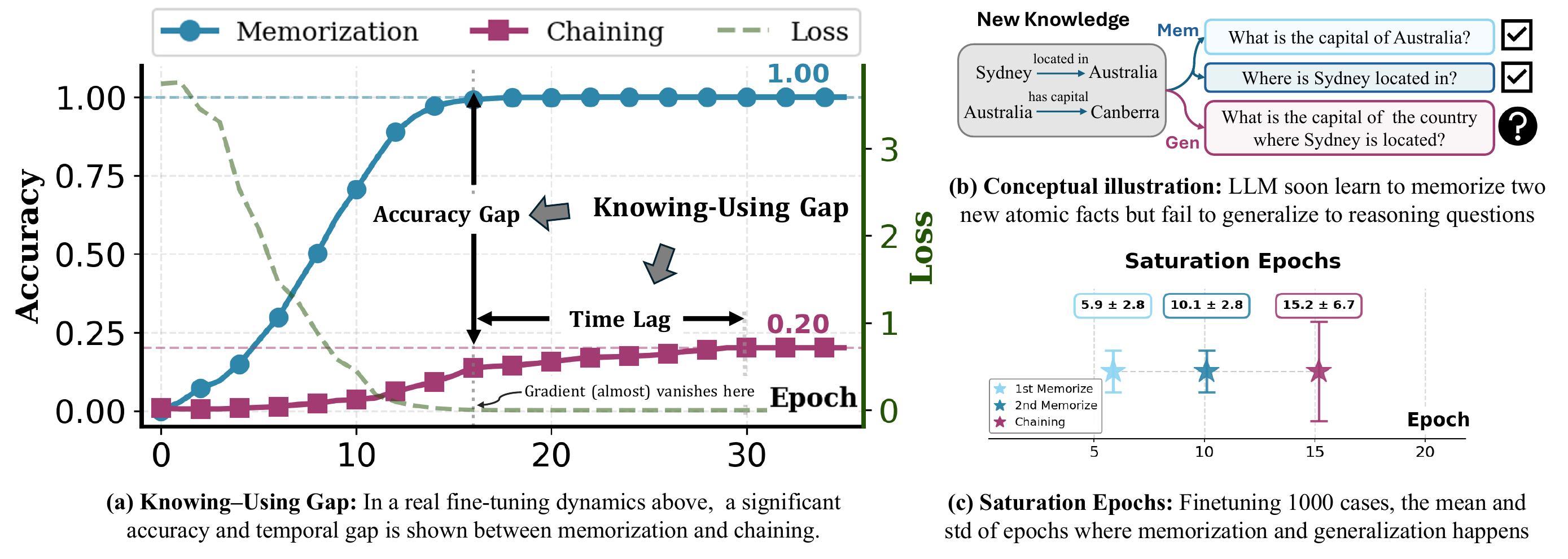}
    \caption{\textbf{Illustration of the Knowing--Using Gap.}}
    \label{fig:knowing_using_gap_curve}
\end{figure*}

Despite sufficient capacity to fit new data \cite{morris2025much,allenphysics33}, LLMs exhibit a ``remembering but not using'' failure \cite{ovadia2024fine, soudani2024fine, zhong2023mquake, cohen2024evaluating, berglundreversal} as shown in Figure~\ref{fig:knowing_using_gap_curve}: models can memorize new facts (e.g., \textit{``Sydney is located in [Entity]''}) but fail to reliably \emph{use} them in downstream reasoning (e.g., \textit{``The capital of the country Sydney located is \dots''}) \cite{zhong2023mquake, cohen2024evaluating, yao2025cake}, creating a gap between simple memorization and flexible generalization.

We term this phenomenon the \textit{Knowing--Using Gap}, characterized by two distinct disparities:
1) an \textbf{accuracy gap}, where generalization accuracy remains significantly lower than memorization; and
2) a \textbf{temporal lag}, where generalization emerges significantly later after remembering.

This observation raises fundamental research questions on the mechanics of fine-tuning: \textit{Once a fact is memorized, when and why does it become accessible to the model's existing reasoning circuits?}

To address these questions, we conduct a fine-grained analysis of the training dynamics during knowledge injection. We construct datasets from two real-world domain knowledge bases and eliminate overlap with pretraining. 
We define two types of reasoning QA tasks to evaluate how LLMs generalize the learned knowledge: \textit{chaining task} requires resolving a bridge entity in the first hop to solve the second; and \textit{intersection task} requires retrieving attributes for two entities and filtering by a required relation. 
These tasks explicitly test whether injected knowledge can be propagated to reasoning beyond recalled in isolation.

To investigate the underlying mechanics, we adapt activation patching~\citep{DBLP:conf/nips/MengBAB22, ghandeharioun2024patchscopes, zhangtowards} into \textit{self-patching}, which copies the hidden representation of an entity from one layer into another layer of the same fine-tuned model and measures whether the correct answer is recovered.
By scanning all layer pairs at every fine-tuning checkpoint, self-patching yields a time-evolving fine-grained spatial map of the permeation of knowledge, identifying which layers and positions contain representations that can unlock the correct answer when routed into an appropriate position.

Based on the observation, we propose the \textit{knowledge--circuit misalignment} hypothesis regarding the Knowing--Using Gap.
Self-patching reveals that after memorization saturates, injected information is retrievable from certain layers but is not reliably integrated into the computation required by multi-hop reasoning. 
Continued fine-tuning after memorization sometimes brings these usable representations into mid-layer computation, coinciding with the emergence of generalization, while in other cases it fails because the representations remain stranded after natural gradient vanishes.
Further interventional experiments provide causal-intervention evidence supporting this hypothesis: by simply relocating memorized representations yields immediate and substantial gains across models and tasks even after natural fine-tuning convergence. 
This suggests that the capability to generalize injected knowledge can be activated artificially, even if it does not naturally emerge during fine-tuning.
Moreover, the gains substantially exceed prompting baselines like CoT and generic perturbation controls, reinforcing the hypothesis that improvements arise from transferring knowledge-relevant information to the reasoning computation path, rather than from superficial decoding effects.
Finally, we turn this diagnosis into a training objective. Layer-wise representation self-distillation (LRSD) aligns, during fine-tuning, the head-entity representation in a middle computation layer with that in a deeper storage layer. With one layer pair fixed per model and no test-time intervention, LRSD improves chaining accuracy in all four models tested and nearly doubles it on Qwen2.5 (\S\ref{sec:lrsd}), moving the contribution from pure diagnosis toward a practical remedy.
To conclude:

\begin{itemize}
  \item We identify and quantify the \textbf{Knowing--Using Gap} during LLM fine-tuning.
  \vspace{-3pt}
  \item We design a \textbf{self-patching} scan that traces \textbf{knowledge permeation} during fine-tuning, locating where the knowledge becomes \emph{causally usable} throughout fine-tuning steps.
  \vspace{-3pt}
  \item We propose the \textbf{knowledge--circuit misalignment hypothesis} and provide mechanistic evidence that manually relocating memorized representations recovers generalization across domains and architectures.
  \vspace{-3pt}
  \item We turn the diagnosis into \textbf{layer-wise representation self-distillation (LRSD)}, a fine-tuning method that aligns storage and computation layer representation to improves generalization, proving its potential to guide the design of generalizable knowledge injection methods.
  \vspace{-3pt}
  \item We release a specialized \textbf{Memorization-to-Generalization dataset} for evaluating multi-hop reasoning on injected knowledge.
\end{itemize}

\definecolor{roleQuery}{RGB}{0,94,160}
\definecolor{roleBridge}{RGB}{204,92,0}
\definecolor{roleAnswer}{RGB}{0,128,88}
\definecolor{roleDistr}{RGB}{135,135,135}
\newcommand{\qent}[1]{\textcolor{roleQuery}{#1}}
\newcommand{\bent}[1]{\textcolor{roleBridge}{#1}}
\newcommand{\aent}[1]{\textcolor{roleAnswer}{#1}}
\newcommand{\ansto}[1]{\,{\color{black!45}$\rightarrow$}\,\textbf{#1}}
\tikzset{
  roleNode/.style={circle, inner sep=0pt, minimum size=4.8pt},
  roleEdge/.style={line width=0.6pt, draw=black!55, -{Stealth[length=1.1mm]}},
}
\newcommand{\taskIconChaining}{%
  \begin{tikzpicture}[baseline=-0.7ex, x=0.46cm]
    \node[roleNode, fill=roleQuery]  (q) at (0,0) {};
    \node[roleNode, fill=roleBridge] (b) at (1,0) {};
    \node[roleNode, fill=roleAnswer] (a) at (2,0) {};
    \draw[roleEdge] (q) -- (b);
    \draw[roleEdge] (b) -- (a);
  \end{tikzpicture}}
\newcommand{\taskIconIntersection}{%
  \begin{tikzpicture}[baseline=-0.7ex, x=0.46cm]
    \node[roleNode, draw=roleDistr, line width=0.7pt] (d) at (-1,0) {};
    \node[roleNode, fill=roleQuery]  (q1) at (0,0) {};
    \node[roleNode, fill=roleAnswer] (a)  at (1,0) {};
    \node[roleNode, fill=roleQuery]  (q2) at (2,0) {};
    \draw[roleEdge] (q1) -- (a);
    \draw[roleEdge] (q2) -- (a);
    \draw[roleEdge, draw=roleDistr, dash pattern=on 1.1pt off 0.9pt] (q1) -- (d);
  \end{tikzpicture}}

\begin{table*}[t]
\centering
\footnotesize
\setlength{\tabcolsep}{4.5pt}
\renewcommand{\arraystretch}{1.18}
\begin{tabularx}{\textwidth}{@{}l l >{\raggedright\arraybackslash}X l@{}}
\toprule
\textbf{Task} & \textbf{Split} & \textbf{Question} & \textbf{Answer} \\
\midrule
\textbf{Chaining} & Train & Which gene or protein is expressed in \qent{embryo}? & \bent{\textbf{IGFBP3}} \\
\taskIconChaining &       & Which drug targets the gene or protein \bent{IGFBP3}? & \aent{\textbf{Mecasermin}} \\
                  & Test  & Which drug targets the genes or proteins that are expressed in \qent{embryo}? & \aent{\textbf{Mecasermin}} \\
\midrule
\textbf{Intersection} & Train & Which gene or protein acts as a carrier for the drug \qent{Acitretin}? & \aent{\textbf{ALB}} \\
\taskIconIntersection &       & Which gene or protein acts as a carrier for the drug \qent{Propiolactone}? & \aent{\textbf{ALB}} \\
                      &       & \leavevmode{\color{roleDistr}What is a side effect of the drug Acitretin?\enspace\textit{(distractor)}} & {\color{roleDistr}\textbf{Tinnitus}} \\
                      & Test  & What gene or protein is a carrier for both \qent{Acitretin} and \qent{Propiolactone}? & \aent{\textbf{ALB}} \\
\bottomrule
\end{tabularx}

\vspace{7pt}
\setlength{\tabcolsep}{4pt}
\renewcommand{\arraystretch}{1.12}
\begin{tabular*}{\textwidth}{@{\extracolsep{\fill}} l l cc cc c @{}}
\toprule
  & & \multicolumn{2}{c}{\textbf{Knowledge graph}} & \multicolumn{2}{c}{\textbf{\# Tasks}} & \textbf{Prior recall} \\
\cmidrule(lr){3-4} \cmidrule(lr){5-6}
\textbf{Source} & \textbf{Domain} & Entities & Relations & Chaining & Intersection &  \\
\midrule
\textsc{STaRK-Prime} & Biomedical & 22{,}193 & 50{,}653 & 11{,}997 & 10{,}000 & 3.8--4.9\% \\
\textsc{STaRK-MAG}   & Academic   & 55{,}149 & 50{,}092 & 12{,}000 & 4{,}390  & 5.0--6.0\% \\
\bottomrule
\end{tabular*}
\caption{\textbf{Tasks and knowledge sources.}
\emph{Top:} verbatim \textsc{STaRK-Prime} examples.
Supporting facts are injected as memorization QA pairs (Train); the generalization question (Test) is held out.
Chaining composes two facts via a bridge entity; intersection finds the entity shared by two facts, ignoring distractors.
Colors mark entity roles: \qent{query}, \bent{bridge}, \aent{answer}, {\color{roleDistr}distractor}.
\emph{Bottom:} task and fact statistics; prior recall (memorization accuracy, range over six base models; per model in Table~\ref{tab:prior_knowledge}) is low, confirming the facts are novel and ruling out knowledge leakage from pre-training.}
\label{tab:data_sample}
\end{table*}

\section{The Knowing--Using Gap}
\label{sec:knowing_using_gap}

\subsection{Data Construction and Experimental Setting}
\label{sec:setting}

To investigate the dynamics of memorization and generalization in LLMs, 
we construct a dataset comprising diverse pairs of memorization and generalization QA tasks. 
Memorization QA tasks serve as the finetuning materials for LLM to memorize new knowledge, while generalization QA tasks, which are not explicitly memorized, test LLM's ability to apply newly acquired knowledge.
The dataset is adapted from STaRK~\cite{DBLP:conf/nips/WuZYHCHISZL24}, 
a real-world semi-structured knowledge base comprising millions of entities and relations of diverse heterogeneous types.
We use the biomedical (STaRK-Prime) and academic (STaRK-MAG) set to ensure the robustness of our findings across domains.
The generation pipeline is detailed in Appendix~\ref{sec:appendix_data}.

We define two types of generalization QA tasks in table~\ref{tab:data_sample} to evaluate how LLMs generalize the learned knowledge in various scenarios.
The atom unit of knowledge in our setting is defined as a fact triplet: $f = (n_{1}, e_{12}, n_{2})$ where $n_{1}, n_{2} \in N$ represent the head and tail entities and $e_{12}$ the relation between them. For example, fact triplet (MRE11, ppi, ATRX) indicates the fact "protein MRE11 interacts with protein ATRX".

\noindent\textbf{Memorization Task.}
For each $f$ a corresponding memorization QA task $mem_{f}$ is generated as the material to memorize the knowledge. Specifically, the task presents the head entity $n_{1}$ and relation $e_{12}$ as the query, 
requiring the model to predict the tail entity $n_{2}$ as the correct answer. 
E.g., $mem_{f}$ = \{Q: "What protein interacts ($e_{12}$) with MRE11 ($n_{1}$)?", A: "ATRX ($n_{2}$)."\}

\noindent\textbf{Generalization Task.}
We design two types of generalization QA tasks in table~\ref{tab:data_sample} from comprehensive meta paths:
\textit{(1) Chaining Task.} This task depends on two supporting tasks on which the model is required to perform sequential reasoning to derive the final answer.
This assesses the model's chain reasoning capabilities.
\textit{(2) Intersection Task.} This task depends on multiple supporting tasks and requires the model to identify shared entities with specific relations out of noise confounders. This tests the model's ability to perform intersection in its knowledge set.

\paragraph{Knowledge Novelty Validation.}
To ensure the injected knowledge is genuinely novel, we first verified that pre-trained models score around or below 6\% accuracy on 1{,}000 randomly sampled memorization tasks before any fine-tuning, on both \textsc{STaRK-Prime} and \textsc{STaRK-MAG} (Table~\ref{tab:prior_knowledge}). Furthermore, during our patching experiments (\S\ref{sec:mechanisms}), we explicitly filter out instances where the base model can already answer the memorization or generalization questions, ruling out potential data leakage.

\noindent\textbf{Evaluation separation.}
Memorization and generalization tasks are disjoint by construction: memorization tasks present single-hop \texttt{(entity, relation $\to$ entity)} queries, while generalization tasks require multi-hop reasoning over \emph{different} query templates unseen in training.
Only fact entities may overlap because chaining inherently requires shared bridge entities, but the compositional query structure is always novel.

\subsection{Formalizing and Characterizing the Gap}
\label{sec:lag_across_tasks}
Let $\theta_t$ denote model parameters after $t$ training steps (or epochs) of knowledge injection on a set of injected facts $\mathcal{K}=\{f_i\}_{i=1}^n$.
We evaluate two time-dependent performance curves.
\textit{Memorization accuracy} $A_{\text{mem}}(t)$: performance on direct recall queries for injected facts (e.g., single-hop completion of a trained triple), and
\textit{Generalization (use) accuracy} $A_{\text{gen}}(t;\mathcal{T})$: performance on a downstream task $\mathcal{T}$ that requires reasoning with injected facts.

We characterize the Knowing--Using Gap along two complementary dimensions:

\noindent\textbf{Accuracy gap.}
At the end of training $t=T_{\max}$, define the difference after convergence as
\begin{equation}
\Delta A(\mathcal{T}) \;=\; A_{\text{mem}}(T_{\max}) \;-\; A_{\text{gen}}(T_{\max};\mathcal{T}).
\end{equation}

\paragraph{Temporal lag.}
To ensure robustness against training fluctuations, we define the saturation time as the earliest point where performance remains stable for at least $w$ consecutive epochs. Specifically, the saturation time of task $\mathcal{T}$ is defined as:
\begin{equation}
T_{\text{gen}}(\mathcal{T}) = \min \{ t : A_{\text{gen}}(t';\mathcal{T}) = 1, \forall t' \in [t, t+w] \}.
\end{equation}
The temporal lag is then defined as $\Delta T(\mathcal{T}) = T_{\text{gen}}(\mathcal{T}) - T_{\text{mem}}$. We exclude failed cases to ensure $T_{\text{gen}}(\mathcal{T})$ reflects the point where generalizable facts are reliably mastered.

\begin{figure}[t]
  \centering
  \includegraphics[width=\textwidth]{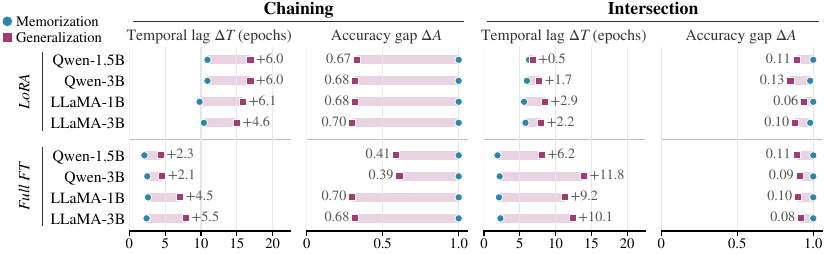}
  \vspace{-4pt}
  \caption{\textbf{The Knowing--Using Gap across tasks, fine-tuning schemes, and models}.
  Circles mark memorization, squares generalization; each bar and its value give the gap.
  $\Delta T = T_{\text{gen}} - T_{\text{mem}}$: epochs until all injected facts are memorized vs.\ generalized.
  $\Delta A = A_{\text{mem}} - A_{\text{gen}}$ at the final epoch.
  Each fact is injected in its own 30-epoch run; values average over runs (exact values in Table~\ref{tab:knowing_using_gap}).}
  \label{fig:knowing_using_gap}
\end{figure}

Figure~\ref{fig:knowing_using_gap_curve} illustrates the Knowing--Using Gap under chaining:
memorization reaches near-ceiling accuracy within a few epochs, while downstream use stays low for an extended period.

Figure~\ref{fig:knowing_using_gap} shows the ubiquity of this dissociation: memorization saturates early, while reliable downstream use is delayed and can remain substantially lower at convergence.
\emph{Across tasks}, chaining shows a two-dimensional gap: it requires substantially more training to become usable and still converges to a low use accuracy.
Intersection is close to an ``ideal'' regime, with high use accuracy and, under LoRA, minimal lag, since its answer often aligns with the memorization answer.
\emph{Across fine-tuning schemes}, full fine-tuning memorizes substantially faster than LoRA, but this advantage does not consistently translate into faster or better downstream use:
chaining still lags and stays at or below 0.61 accuracy, and intersection shows a markedly larger lag.
\emph{Across scales}, increasing model size does not eliminate the gap, and more injected facts tend to widen $\Delta A$ (Appendix~\ref{sec:appendix_scaling}).

\WFclear
\section{Mechanistic Analysis for Knowledge--Circuit Misalignment}
\label{sec:mechanisms}

\begin{figure*}[ht]
    \centering
    \includegraphics[width=0.9\linewidth]{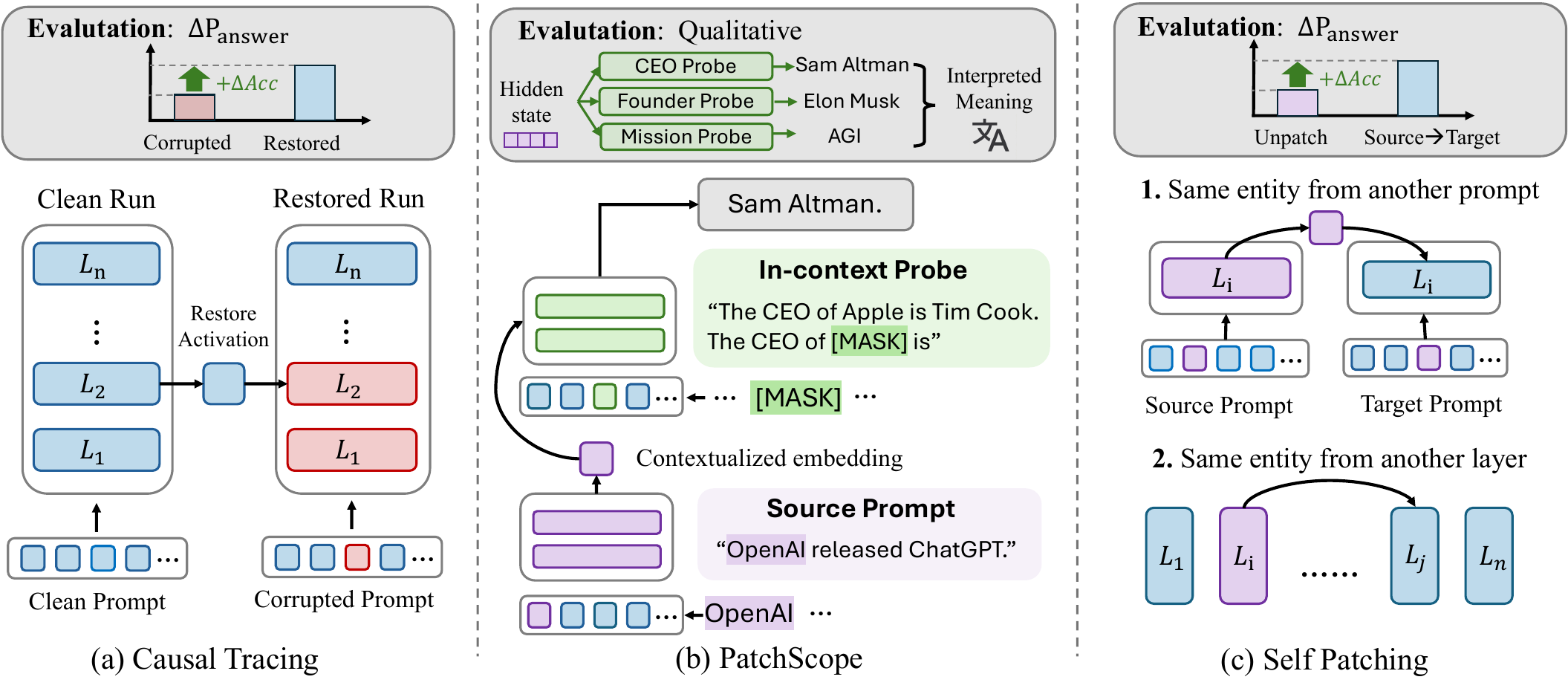}
    \vspace{-1em}
    \caption{\textbf{Method Comparison.}
    \emph{Causal tracing} corrupts a run to identify causally relevant locations.
    \emph{Patchscope} interprets a hidden state by decoding it into natural language.
    \emph{Self-patching} tests the effect position by swapping its internal representation across layers and contexts.}
    \vspace{-1em}
    \label{fig:methods}
\end{figure*}

This section investigates the mechanism underlying the Knowing--Using Gap.
We hypothesize that injected facts reside in hidden states but are not routed through the layers performing multi-hop reasoning, and show that relocating them substantially restores generalization.
\S\ref{sec:lrsd} turns this diagnosis into a fine-tuning objective.

\noindent\textbf{Knowledge--circuit misalignment hypothesis.}
A Knowing--Using Gap is driven by a spatial misalignment of knowledge storage and reasoning circuits: fine-tuning first encodes new facts in easy-to-fit storage states for memorization (often early or very late layers) that support direct recall, but are not reliably routed into the effective positions that are causally required for multi-step reasoning, often in mid-layer computation as shown in previous literature and our experiment in Figure~\ref{fig:patching_position_distribution}.

This hypothesis predicts that (1) after memorization saturates, there should exist \emph{off-path} representations that already contain injected information but do not yield correct reasoning end-to-end; and (2) explicitly relocating those representations into effective locations should immediately increase downstream use.
We test these predictions with several causal intervention tools, mainly self-patching.

\vspace{-1em}
\subsection{Tracing Knowledge Permeation with Self-Patching}
\label{section:method}

\begin{wrapfigure}{r}{0.53\textwidth}
  \vspace{-1.6em}
  \refstepcounter{algorithm}\label{alg:self_patching}
  \begingroup
  \noindent\begin{minipage}{0.96\linewidth}
    \hrule height 0.65pt
    \vspace{0.35em}
    \textbf{Algorithm~\thealgorithm: Self-patching scan}
    \vspace{0.25em}
    \hrule height 0.35pt
    \vspace{0.35em}
    \textbf{Input:} model $M$, prompts $P_s,P_t$, anchor $E$, answer $y^*$, score $I$.\\
    \textbf{For} each layer pair $(l_{\text{src}},l_{\text{tgt}})$:
    \begin{itemize}[label=$\triangleright$,leftmargin=1.25em,labelsep=0.35em,topsep=1pt,itemsep=1pt,parsep=0pt]
      \item Locate anchor tokens $T_s=T(P_s,E)$ and $T_t=T(P_t,E)$.
      \item Cache source state $z=h^{l_{\text{src}}}_{T_s}(P_s)$.
      \item Run $P_t$ and replace $\tilde{h}^{l_{\text{tgt}}}_{T_t}\leftarrow z$.
      \item Continue the forward pass to obtain $\tilde{M}(P_t)$.
      \item Record $\Delta I=I(\tilde{M}(P_t),y^*)-I(M(P_t),y^*)$.
    \end{itemize}
    \textbf{Output:} layer-pair map $A[l_{\text{src}},l_{\text{tgt}}]=\Delta I$.
    \vspace{0.35em}
    \hrule height 0.65pt
  \end{minipage}
  \endgroup
  \vspace{-1.2em}
\end{wrapfigure}

Self-patching is a form of activation patching~\citep{DBLP:conf/nips/MengBAB22,zhangtowards,ghandeharioun2024patchscopes}, adapted here to trace knowledge permeation (Algorithm~\ref{alg:self_patching}).
It copies the residual-stream states $h^{l_{\text{src}}}_{T_s}(P_s)$ of an $L$-layer model $M$ at layer $l_{\text{src}}$ over the tokens $T_s=T(P_s,E)$ of the head entity $E$ in a source prompt $P_s$ into layer $l_{\text{tgt}}$ of a target prompt $P_t$; $I$ scores the resumed run $\tilde{M}(P_t)$ against the answer $y^*$ by exact match or mean reciprocal rank.

\begin{figure*}[ht]
    \centering
    \includegraphics[width=\linewidth]{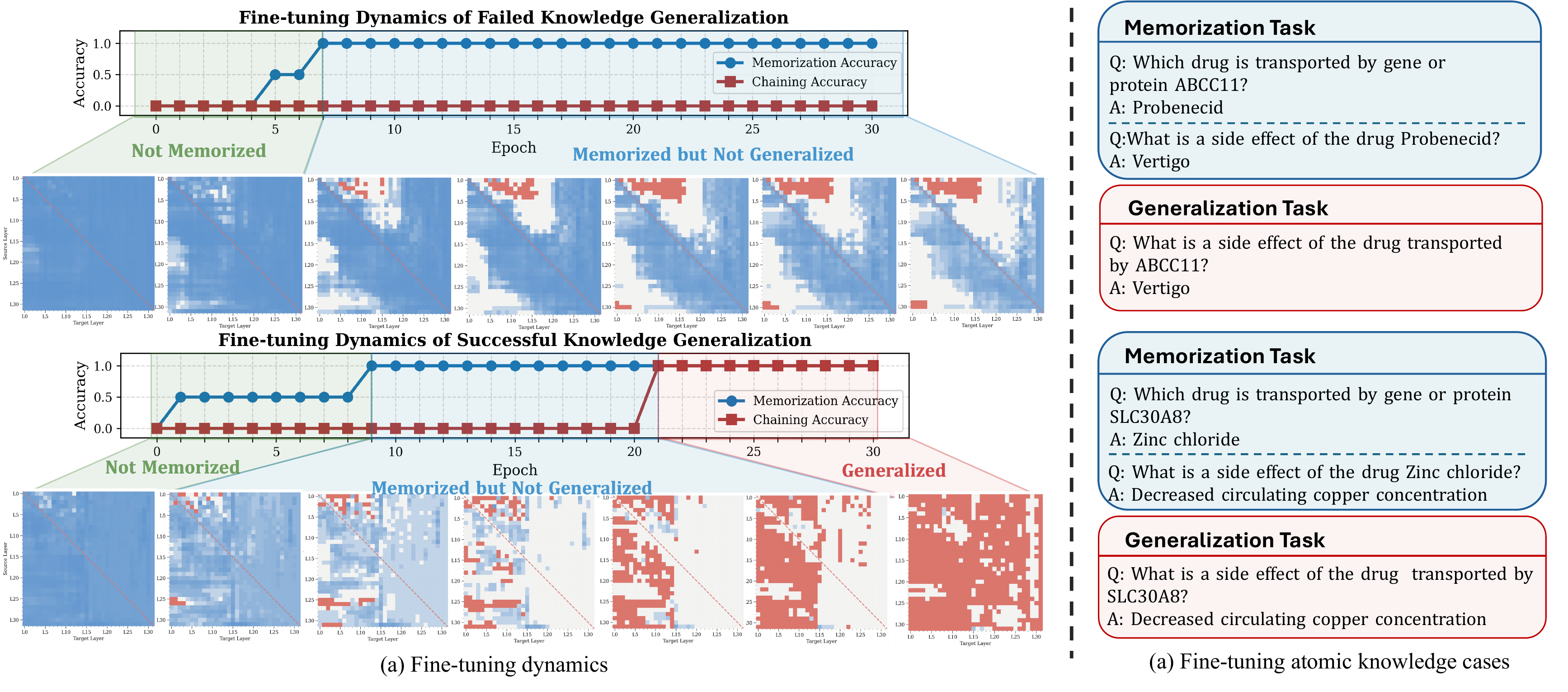}
    \vspace{-1em}
    \caption{\textbf{Permeation dynamics of knowledge aligned with fine-tuning process.}
    Cell $(l_{\text{src}}, l_{\text{tgt}})$ measures whether patching the head-entity representation from $l_{\text{src}}$ to $l_{\text{tgt}}$ increases generalization accuracy, with red indicating full recovery, blue for no effect, and white for partial gain.
    \textbf{Top:} a natural failed instance. The red region indicates the position of knowledge storage, which permeates but halts before reaching the diagonal, indicating generalization failure. 
    \textbf{Bottom:} a successful instance, patch-effective regions expand and cover diagonal before training stabilizes.}
    \vspace{-1em}
    \label{fig:chaining_dynamics}
\end{figure*}

\noindent\textbf{Scanning layer pairs across training.}
Patching a multi-hop query onto itself ($P_s=P_t=P_{\text{gen}}$) only relocates the model's own representation and, unlike causal tracing (Figure~\ref{fig:methods}), needs no correct run, so it applies to failed cases.
In the $L\times L$ map of patched scores $I(\tilde{M}(P_t),y^*)$, the diagonal ($l_{\text{src}}=l_{\text{tgt}}$) is the unpatched model, and a high off-diagonal cell marks a representation at $l_{\text{src}}$ usable once placed at $l_{\text{tgt}}$.
Repeated each epoch, the maps trace permeation as high-score cells spreading toward the diagonal, and generalization as their arrival (Figure~\ref{fig:chaining_dynamics}).

\subsection{Knowledge Dynamics in Fine-Tuning}
\label{sec:permeation}

We first trace the knowledge permeation of two atomic knowledge during fine-tuning to understand \emph{after a fact is memorized, when, where and how does it become accessible to the model's pre-existing reasoning logic.}
We save checkpoints each epoch and perform layer-to-layer self-patching at the head-entity positions $E_{head}$ on $P_{\text{gen}}$, producing a heatmap over $(l_{\text{src}}, l_{\text{tgt}})$ at each step.

Before memorization, patching rarely helps as indicated by the all-blue map: injected knowledge is not yet available internally at all, thus swapping can not help.

On the moment in training curve when two facts are memorized, we can observe clear \emph{off-diagonal} red region appear in Figure~\ref{fig:chaining_dynamics} that says the needed information is already stored in certain layers and can take effect if routed into particular target layers. 
However, the model still fails end-to-end in the natural fine-tuning process: diagonal cells remains blue (fail with no intervention).

As fine-tuning continues, the red region expands wider towards the heatmap diagonal, indicating the gradual permeation of knowledge into more layers.
In successful cases, the red region covers the diagonal line, which means the natural emergence of generalization from fine-tuning.
In failed cases, while still expanding, the red region halts before reaching the diagonal. This is due to the fine-tuning nature that after memorization where $M(P)$ matches $y^{*}$, loss will diminish soon and become too small to drive further gradient updates for internal change. 
The model will stuck and fail to generalize unless manually relocating the representation, as shown in the next section.
This knowledge permeation dynamics aligns well with the training phases in time, which validates the prediction from our hypothesis: knowledge is stored but not used until the representation becomes available in locations effective for the reasoning computation.

\begin{table*}[ht]
  \centering
  \small
  \setlength{\tabcolsep}{4.5pt}
  \renewcommand{\arraystretch}{1.14}
  \begin{tabular*}{\linewidth}{@{\extracolsep{\fill}} l c c c c c c c c c c @{}}
    \toprule
    \textbf{Model}
      & \multicolumn{5}{c}{\makecell[c]{\textbf{STaRK-Prime}}}
      & \multicolumn{5}{c}{\makecell[c]{\textbf{STaRK-MAG}}} \\
    \cmidrule(lr){2-6} \cmidrule(lr){7-11}
      & \textbf{Mem.} & \multicolumn{2}{c}{\textbf{Chaining}} & \multicolumn{2}{c}{\textbf{Intersection}}
      & \textbf{Mem.} & \multicolumn{2}{c}{\textbf{Chaining}} & \multicolumn{2}{c}{\textbf{Intersection}} \\
    \cmidrule(lr){3-4} \cmidrule(lr){5-6} \cmidrule(lr){8-9} \cmidrule(lr){10-11}
      &  & \textbf{w/o} & \cellcolor{gray!15}\textbf{pat.} & \textbf{w/o} & \cellcolor{gray!15}\textbf{pat.}
      &  & \textbf{w/o} & \cellcolor{gray!15}\textbf{pat.} & \textbf{w/o} & \cellcolor{gray!15}\textbf{pat.} \\
    \midrule
    Qwen-2.5-1.5B & 0.998 & 0.078 & \cellcolor{gray!15}0.440 & 0.793 & \cellcolor{gray!15}0.987 & 0.991 & 0.046 & \cellcolor{gray!15}0.286 & 0.928 & \cellcolor{gray!15}0.994 \\
    Qwen-2.5-3B   & 0.997 & 0.114 & \cellcolor{gray!15}0.542 & 0.798 & \cellcolor{gray!15}0.986 & 0.986 & 0.052 & \cellcolor{gray!15}0.288 & 0.958 & \cellcolor{gray!15}0.994 \\
    Qwen-2.5-7B   & 0.996 & 0.124 & \cellcolor{gray!15}0.504 & 0.774 & \cellcolor{gray!15}0.956 & 0.982 & 0.068 & \cellcolor{gray!15}0.310 & 0.976 & \cellcolor{gray!15}0.994 \\
    \midrule
    LLaMA-3.2-1B  & 0.994 & 0.102 & \cellcolor{gray!15}0.316 & 0.874 & \cellcolor{gray!15}0.975 & 0.975 & 0.082 & \cellcolor{gray!15}0.182 & 0.982 & \cellcolor{gray!15}0.994 \\
    LLaMA-3.2-3B  & 0.993 & 0.126 & \cellcolor{gray!15}0.404 & 0.815 & \cellcolor{gray!15}0.969 & 0.974 & 0.064 & \cellcolor{gray!15}0.240 & 0.982 & \cellcolor{gray!15}1.000 \\
    LLaMA-3.1-8B  & 0.986 & 0.182 & \cellcolor{gray!15}0.458 & 0.795 & \cellcolor{gray!15}0.921 & 0.975 & 0.072 & \cellcolor{gray!15}0.222 & 0.982 & \cellcolor{gray!15}0.994 \\
    \bottomrule
  \end{tabular*}
  \caption{\textbf{Oracle self-patching recovers downstream use across models, architectures, and knowledge domains.}
  ``pat.'' denotes head-entity self-patching at the most effective layer, averaged over 1000 injected facts.}
  \label{tab:patching_results}
  \vspace{-2em}
\end{table*}

\subsection{Manually Trigger Generalization}
\label{sec:manual_alignment}
We then quantify the \emph{oracle upper bound} at convergence to see how much downstream use can be activated by relocating representations.
Since self-patch do not introduce new information but only relocates existing representations, the substantial improvement in downstream use after patching indicates that the generalization failure is not due to the absence of knowledge but rather a misalignment between storage and computation.
We systematically fine-tune on different architectures, scales, and knowledge domains, and perform self-patching at convergence to measure the recoverable headroom.
Experiments are computed with 1,000 samples to avoid randomness, detailing in Appendix~\ref{sec:appendix_experiment_setup}.    
Table~\ref{tab:patching_results} shows that memorization saturates near-perfectly across models and domains, yet downstream use without intervention remains low, especially for chaining.
Oracle self-patching  yields large and consistent improvements along all architectures, model scales and knowledge domains.
Chaining accuracy lifts by 2.2--6.2$\times$ in every cell, and the smaller know-use gap in intersection is nearly eliminated.
The uniform improvement again points to a structural origin rather than an random artifact, showing causal evidence for the knowledge--circuit misalignment hypothesis.
\label{sec:cross_domain}

\begin{figure*}[ht]
    \centering
    \includegraphics[width=\linewidth]{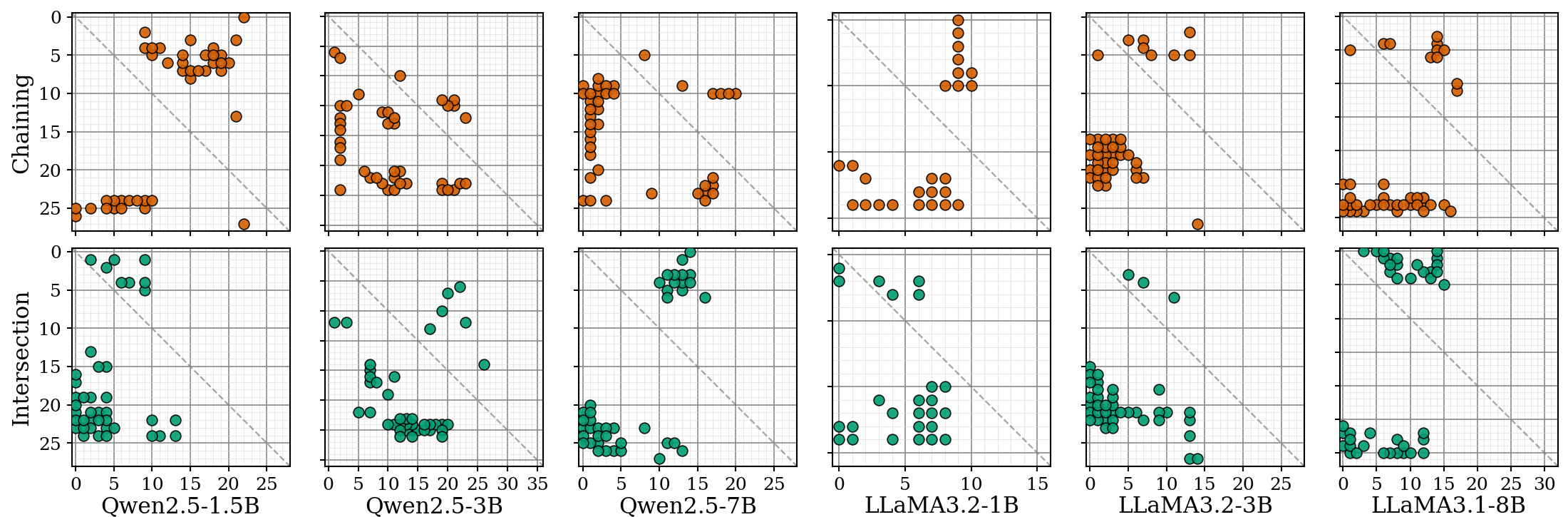}
    \vspace{-2em} %
    \caption{\textbf{Effective patch locations concentrate into two clusters.}
    Dots means the top-40 layer pairs that recover most generalizaiton failure on the whole dataset; $y$-axis is $l_{\text{src}}$, $x$-axis is $l_{\text{tgt}}$.}
    \label{fig:patching_position_distribution}
\end{figure*}

\vspace{-1em}
\noindent\textbf{Location of most effective patching.}
\label{sec:where_effective}
In Figure~\ref{fig:patching_position_distribution}, the most effective source layers form an early and a late cluster, both patching into middle target layers.
The late cluster is intuitive~\citep{ghandeharioun2024patchscopes,biran2024hopping}, as later representations carry enriched information; the early one is surprising: moved into middle layers, early representations also trigger generalization, even skipping several layers.
Hence the needed information does not emerge too late: it is stored in both early and late layers but fails to align with the mid-layer computation that bottlenecks its use.
Late target layers are useless (empty right region), as the reasoning stream has moved to the last token there~\citep{geva2023dissecting}.
This distribution also informs the design and layer choice of LRSD (\S\ref{sec:lrsd}).

\subsection{Ablations and Sanity Checks}
\label{sec:controls_prompting}
\phantomsection
\label{sec:token_ablation}
\begin{wraptable}{r}{0.45\textwidth}
  \vspace{-2em}
  \centering
  \caption{\textbf{Token-position ablation.}
  McNemar's test on patching at different positions against random baseline.}
  \label{tab:mcnemar_position}
  \footnotesize
  \resizebox{\linewidth}{!}{%
    \begin{tabular}{lccc}
      \toprule
      \textbf{Position} & {\textbf{Mean}} & \textbf{$p$-value} & \textbf{Significance} \\
      \midrule
      \textbf{Random}          & 0.1359 & --      & --  \\
      \midrule
      \textbf{<BOS>}           & 0.0485 & 1.0000  & n.s. \\
      \textbf{<EOS>}           & 0.4029 & 0.0000  & *** \\
      \textbf{First Relation}  & 0.1990 & 0.0235  & * \\
      \textbf{Second Relation} & 0.1990 & 0.0121  & * \\
      \rowcolor{gray!15}
      \textbf{Entity}          & 0.6408 & 0.0000  & *** \\
      \bottomrule
    \end{tabular}%
  }
  \vspace{-1em}
\end{wraptable}
We perform several ablations to rule out other potential explanations such as position, prompting and perturbation artifacts.

\noindent\textbf{Token-position Ablations.}
We experimentally prove that injected factual information is tightly tied to entity mentions but not position-agonistic.
Table~\ref{tab:mcnemar_position} shows that patching at the head-entity position yields the largest and most significant improvements, far exceeding random and <BOS>.
The <EOS> position exhibits a secondary signal, consistent with late positions aggregating information relevant for generation.

\begin{table}[ht]
  \centering
  {%
  \footnotesize
  \setlength{\tabcolsep}{3.5pt}
  \renewcommand{\arraystretch}{1.1}
  \resizebox{\columnwidth}{!}{%
  \begin{tabular}{l
      S[table-format=1.3] S[table-format=1.3] S[table-format=1.3] S[table-format=1.3]
      S[table-format=1.3] S[table-format=1.3] S[table-format=1.3] S[table-format=1.3]}
    \toprule
    & \multicolumn{4}{c}{\textbf{Chaining}} & \multicolumn{4}{c}{\textbf{Intersection}} \\
    \cmidrule(lr){2-5}\cmidrule(lr){6-9}
    {\textbf{Model}}
      & {\makecell[c]{\textbf{w/o pat.}}}
      & {\makecell[c]{\textbf{CoT}}}
      & {\makecell[c]{\textbf{Irrelevant pat.}}}
      & {\makecell[c]{\textbf{Self pat.}}}
      & {\makecell[c]{\textbf{w/o pat.}}}
      & {\makecell[c]{\textbf{CoT}}}
      & {\makecell[c]{\textbf{Irrelevant pat.}}}
      & {\makecell[c]{\textbf{Self pat.}}} \\
    \midrule
    Qwen-2.5-1.5B & 0.078 & 0.132 & 0.150 & 0.440 & 0.793 & 0.243 & 0.873 & 0.987 \\
    Qwen-2.5-3B   & 0.114 & 0.288 & 0.184 & 0.542 & 0.798 & 0.387 & 0.856 & 0.986 \\
    Qwen-2.5-7B   & 0.124 & 0.390 & 0.194 & 0.504 & 0.774 & 0.398 & 0.890 & 0.956 \\
    \bottomrule
  \end{tabular}%
  }
  }
  \caption{\textbf{Prompting and perturbation ablations.}
  Direct/no patching and self-patching are shared anchors; CoT tests prompt effects, and irrelevant patching controls for generic activation perturbations.}
  \vspace{-1em} %
  \label{tab:prompting_patch_controls}
\end{table}
\WFclear

\noindent\textbf{Control experiment on prompting and perturbation.}
Table~\ref{tab:prompting_patch_controls} combines two complementary ablations on prompting and representation.
CoT prompting improves chaining but remains far below self-patching, and sometimes even degrades intersection performance, and irrelevant patching that uses representation from unrelated fact still lags largely behind patching \emph{correct} fact's representation.

\noindent\textbf{Detect knowledge existence}. To rule out the chance that gains are driven by prompt artifacts or noises, we further cross-context patch the $E_{head}$ representation from the single-hop memorization query $P_\text{mem}$ which contains knowledge representation to $P_\text{gen}$.
Figure~\ref{fig:mem_self_patching_comparison} (Appendix~\ref{sec:appendix_misalignment}) shows strong correlation in the effective layer-pair patterns from both contexts, supporting that self-patching transfers injected-knowledge representations similar to $P_\text{mem}$ rather than some unknown noises.

\WFclear

\subsection{Layer-wise Self-Distillation}
\label{sec:lrsd}

Based on the finding that aligning the late layers where knowledge is stored with the middle computation layers facilitates generalization (Figure~\ref{fig:patching_position_distribution}), 
while plain fine-tuning stops driving knowledge permeation after memorization(\S\ref{sec:permeation}), we build the alignment into training to improve generalization.

\noindent\textbf{Method.}
Layer-wise representation self-distillation (LRSD) aligns the head-entity representation at a middle computation layer $l_{\text{tgt}}$ with the one at a late storage layer $l_{\text{src}}>l_{\text{tgt}}$, within the same forward pass on each memorization prompt $P$.
It is the training-time counterpart of patching from $l_{\text{src}}$ into $l_{\text{tgt}}$, and a form of hidden-state self-distillation~\citep{romero2015fitnets,zhang2019byot} in which the deeper layer acts as the teacher.
Let $T=T(P,E_{\text{head}})$ be the head-entity tokens, $h^{l}_{i}$ the state of token $i$ after layer $l$, and $y^*$ the answer; fine-tuning then minimizes
\begin{equation}
\label{eq:lrsd}
\mathcal{L}_{\text{LRSD}}(P)=\frac{1}{|T|}\sum_{i\in T}\frac{\bigl\lVert h^{l_{\text{tgt}}}_{i}-\operatorname{sg}\bigl(h^{l_{\text{src}}}_{i}\bigr)\bigr\rVert_2^2}{\bigl\lVert h^{l_{\text{src}}}_{i}\bigr\rVert_2^2},
\qquad
\mathcal{L}=\mathcal{L}_{\text{CE}}(y^*\mid P)+\lambda\,\mathcal{L}_{\text{LRSD}}(P).
\end{equation}
The stop-gradient $\operatorname{sg}$ keeps the teacher fixed, so the alignment gradient only reaches the layers up to $l_{\text{tgt}}$, and dividing by the teacher's norm makes $\lambda$ comparable across layers and models.
LRSD needs no oracle, no multi-hop supervision and no extra forward pass, and the trained model is used as is (Algorithm~\ref{alg:lrsd}).

\begin{wrapfigure}{r}{0.44\textwidth}
  \vspace{-1.3em}
  \refstepcounter{algorithm}\label{alg:lrsd}
  \noindent\begin{minipage}{0.98\linewidth}
    \raggedright
    \hrule height 0.65pt
    \vspace{0.35em}
    \textbf{Algorithm~\thealgorithm: LRSD fine-tuning}
    \vspace{0.25em}
    \hrule height 0.35pt
    \vspace{0.35em}
    \textbf{Input:} $M_\theta$, $\mathcal{D}_{\text{mem}}$, layers $l_{\text{src}}>l_{\text{tgt}}$, weight $\lambda$.\\
    \textbf{For} each mini-batch $B\subset\mathcal{D}_{\text{mem}}$:
    \begin{itemize}[label=$\triangleright$,leftmargin=1.25em,labelsep=0.35em,topsep=1pt,itemsep=1pt,parsep=0pt]
      \item Forward $B$, keeping $h^{l_{\text{tgt}}}_{T}$ and $h^{l_{\text{src}}}_{T}$.
      \item $\mathcal{L}_B\leftarrow$ sum of $\mathcal{L}$ over $B$ (Eq.~\ref{eq:lrsd}).
      \item $\theta\leftarrow\theta-\eta\nabla_\theta\mathcal{L}_B$, with $h^{l_{\text{src}}}_{T}$ detached.
    \end{itemize}
    \textbf{Output:} $M_\theta$, used without patching.
    \vspace{0.35em}
    \hrule height 0.65pt
  \end{minipage}
  \vspace{-1.3em}
\end{wrapfigure}

\noindent\textbf{Setup.}
We fine-tune on Qwen2.5-1.5B/3B and LLaMA-3.2-1B/3B and compare LRSD at two loss weights, $\lambda=0.1$ and $1$, against plain fine-tuning over three seeds.
The design follows the diagnosis: LRSD acts on the head entity, the position where patching works best, and pulls the layer at half depth toward the layer at three-quarter depth, a late-to-middle pair from the cluster in Figure~\ref{fig:patching_position_distribution}.
We chose this pair in a small pilot on Qwen2.5-3B and kept the same relative depth for the other models.
The trained models are evaluated without any patching, by exact match on all chaining questions averaged over the last 20 of the 50 epochs; Appendix~\ref{sec:appendix_lrsd} gives the full details and results.

\begin{figure}[t]
  \centering
  \includegraphics[width=\linewidth]{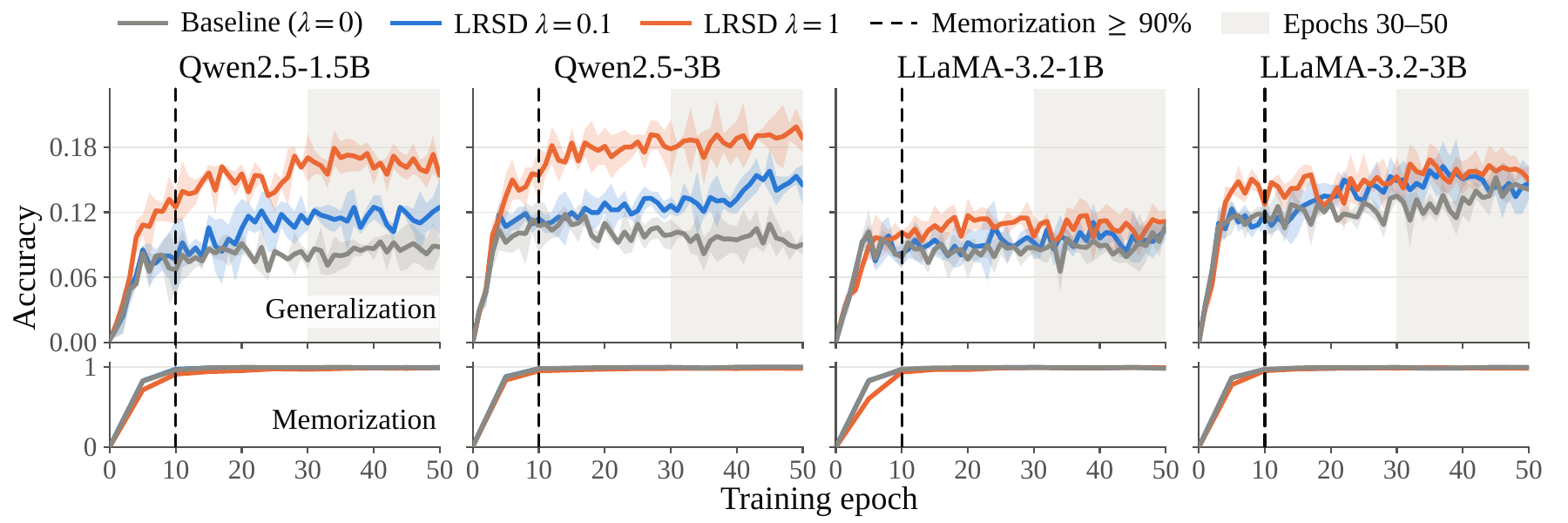}
  \vspace{-1.6em}
  \caption{\textbf{Aligning the late storage layer with the middle layer improves chaining without hurting memorization.}
  Chaining (top) and memorization (bottom) accuracy during fine-tuning, averaged over three seeds with one standard deviation shaded.
  The dashed line marks the epoch by which every run has memorized at least 90\% of the facts, and the grey band is the evaluation window.}
  \label{fig:lrsd_chaining}
  \vspace{-1em}
\end{figure}

\noindent\textbf{Results.}
By aligning the late-layer representation where the knowledge is stored with the middle layer, LRSD improves chaining in every model (Figure~\ref{fig:lrsd_chaining}).
With $\lambda=1$, chaining accuracy nearly doubles on both Qwen2.5 models, a relative gain of 95\%, and rises by about 20\% on both LLaMA-3.2 models; the weaker weight $\lambda=0.1$ still helps every model, while memorization stays intact as plain fine-tuning.
It keeps growing after memorization saturates, the phase in which plain fine-tuning stays flat, consistent with LRSD continuing the relocation that the vanishing memorization loss no longer drives.
Two controls on Qwen2.5-3B support the design: applying the same loss to the tokens before the entity, or between two late layers, brings little or no gain.

\section{Related Work}

\noindent\textbf{Mechanistic interpretability.} Mechanistic interpretability aims to reverse-engineer neural networks into human-understandable components, moving beyond behavioral analysis to causal explanations of model internals~\cite{zhang2026locate}. Previous methods can be categorized into observation-based methods, such as logit-lens \cite{nostalgebraist2020logitlens, DBLP:conf/acl/WendlerVM024}, linear probes \cite{DBLP:conf/iclr/AlainB17, DBLP:journals/coling/Belinkov22}, and sparse auto-encoders \cite{DBLP:conf/iclr/HubenCRES24, DBLP:conf/iclr/GaoTTGTRSL025} which disentangles the polysemantic features; and intervention-based methods, such as causal tracing \cite{DBLP:conf/nips/MengBAB22, DBLP:conf/iccvw/PalitPAL23} and activation patching \cite{DBLP:conf/iclr/WangVCSS23}. Recent research has shifted from analyze individual neurons to circuits \cite{DBLP:conf/nips/Yao0XWXDC24}, which are subgraphs of the model responsible for specific behaviors. For instance, "induction heads" have been identified as the primary mechanism for in-context learning \cite{DBLP:journals/corr/abs-2209-11895}, while other studies have mapped circuits responsible for indirect object identification \cite{DBLP:conf/iclr/WangVCSS23} and entity tracking \cite{DBLP:conf/iclr/PrakashSHBB24}. 

\noindent\textbf{Knowledge Representation in LLM.} The "Linear Representation Hypothesis" \cite{DBLP:conf/icml/ParkCV24} and the "Key-Value Memory" \cite{DBLP:journals/corr/abs-2501-02950} framework posits that LLMs encode factual knowledge (e.g., "A is B") as linear directions in the activation space, often stored within the weights of MLP layers \cite{DBLP:conf/nips/MengBAB22, DBLP:conf/acl/DaiDHSCW22}. Based on this localization, Model Editing techniques like ROME \cite{DBLP:conf/nips/MengBAB22} were developed to directly update specific facts by modifying MLP weights. However, a critical limitation of these approaches is the gap between storing a fact and utilizing it for reasoning \cite{gupta2024model}. Recent benchmarks like MQuAKE \cite{zhong2023mquake} and RippleEdits \cite{cohen2024evaluating} reveal that while models can recall edited facts (high memorization), they fail to propagate these updates to multi-hop reasoning tasks. 

\noindent\textbf{Grokking and Learning dynamics.} Grokking \cite{power2022grokking, wang2024grokking, DBLP:conf/nips/LiuKNMTW22} refers to the phenomenon that generalization performance on validation sets suddenly improves long after training accuracy has saturated. 
Originally observed in small algorithmic tasks \cite{power2022grokking}, grokking has recently been confirmed in pre-training and fine-tuning of large transformers \cite{DBLP:journals/corr/abs-2506-21551,DBLP:conf/iclr/NandaCLSS23,wang2024grokking}. 

\vspace{-1.5em}
\section{Conclusion}
\label{sec:conclusion}
\vspace{-1em}
We identified the Knowing--Using Gap: fine-tuned LLMs memorize injected facts yet fail to use them in multi-hop reasoning. Self-patching traces this failure to \textit{knowledge--circuit misalignment}: knowledge stored in early or late layers does not permeate into the mid-layer circuits required for reasoning.
The gap is thus a routing problem, not a capacity one, and it is partially reversible: 
relocating the stranded representations recovers much of the lost accuracy. We further proposed a demonstrative methed LRSD, which aligns storage and computation layers during fine-tuning and consistently improves generalization, proving the hypothesis' potential in design practical method.

\subsection*{AI use statement}

In this work, we used generative AI tools for synthetic data generation, research methodology design, implementation, data cleaning, and qualitative analysis.
Specifically, an LLM (Deepseek API) phrased part of the memorization and generalization questions from facts and paths sampled from the knowledge graph (Appendix~\ref{sec:appendix_data}); the facts themselves come from the knowledge graph.
An AI coding assistant implemented the LRSD analysis and visualization scripts, and categorized model errors afterwards. (Section~\ref{sec:lrsd}, Appendix~\ref{sec:appendix_lrsd}); the final experimental designs were chosen by the authors.
Generative AI tools were also used for data cleaning.
The code of the core pipeline, including dataset construction, fine-tuning, and self-patching, was written by the authors.
We have not used generative AI tools for hypothesis refinement, mathematical formulation, translation, or results interpretation, and theoretical modeling and proof development are not applicable to this work.
Additionally, we used generative AI tools for polishing the writing, suggesting the structure of the paper, and writing plotting scripts for figures.
We have reviewed all AI-assisted work: a random sample of the LLM-generated questions was manually checked against the source knowledge-graph facts, and all AI-written code was reviewed and tested by the authors.
We take responsibility for the final content of this work, including text, claims or artifacts produced with the aid of generative AI.

\subsection*{Ethics statement}

The authors reviewed the ICLR Code of Ethics and believe the work conforms to it. The work does not involve crowdsourcing, human-subject experiments, participant compensation, or IRB-style review requirements.
The study adapts the \textsc{STaRK} benchmark~\cite{DBLP:conf/nips/WuZYHCHISZL24}, specifically \textsc{STaRK-Prime} and \textsc{STaRK-MAG}; the public STaRK repository is released under the MIT License.\footnote{\url{https://github.com/snap-stanford/stark}} We use open-weight Qwen2.5 models under their public licenses: Qwen2.5-1.5B and Qwen2.5-7B under Apache 2.0, and Qwen2.5-3B under the Qwen Research/Tongyi Qianwen Research terms.\footnote{\url{https://qwen2.org/qwen2-5/}} We use LLaMA-3.1 and LLaMA-3.2 models under the Meta Llama Community License terms.\footnote{\url{https://huggingface.co/meta-llama/Llama-3.1-405B/blob/main/LICENSE} and \url{https://huggingface.co/meta-llama/Llama-3.2-3B/blob/main/LICENSE.txt}}
We release the derived memorization-to-generalization QA data and accompanying code as new artifacts under the MIT License in our repository. The released data are generated from knowledge-graph facts and templates or AI-assisted wording; no consent-bearing human-subject data are collected for this work.

\subsection*{Reproducibility statement}

Code, data, and reproduction instructions are included in our release at \url{https://anonymous.4open.science/r/Mem2Gen-71FF}.
Appendix~\ref{sec:appendix_data} describes the data construction, Appendix~\ref{sec:appendix_experiment_setup} the injection protocols, hyperparameters, evaluation rules, self-patching procedures and compute, and Appendix~\ref{sec:appendix_lrsd} the layer-wise self-distillation runs.

\bibliography{custom}
\bibliographystyle{iclr2027_conference}

\appendix

\clearpage
\noindent\textbf{Appendix overview.}
The appendices follow the order of the main text.
Appendices~\ref{sec:appendix_data} and~\ref{sec:appendix_experiment_setup} describe the data and the protocols behind all experiments;
Appendix~\ref{sec:appendix_scaling} gives the exact values and scale checks of the Knowing--Using Gap (Section~\ref{sec:lag_across_tasks});
Appendices~\ref{sec:appendix_permeation}, \ref{sec:appendix_mag} and~\ref{sec:appendix_misalignment} extend the mechanistic analysis with more permeation maps (Section~\ref{sec:permeation}), the STaRK-MAG replication (Section~\ref{sec:manual_alignment}) and further controls (Section~\ref{sec:controls_prompting});
Appendix~\ref{sec:appendix_lrsd} details layer-wise self-distillation (Section~\ref{sec:lrsd}), and Appendix~\ref{sec:appendix_responsible} discusses limitations.

\section{Dataset Construction and Validation}
\label{sec:appendix_data}

This appendix details the data of Section~\ref{sec:setting}.

\subsection{Construction Pipeline}
\label{sec:appendix_data_pipeline}

We sample facts from the knowledge graph and turn them into question--answer pairs in three steps:

\begin{enumerate}[leftmargin=1.5em,itemsep=1pt,topsep=2pt]
    \item \textbf{Sample valid paths.} For each type of generalization task,
    we manually design a list of valid meta paths to guide the sampling process and generate meaningful and reasonable questions.
    A meta path defines a sequence of entity types and relation types that outline the structure of the knowledge to be used for the task.
    For instance, a meta-path for chaining task may be denoted as $(\text{anatomy} \to \text{expression present} \to \text{gene/protein} \to \text{target} \to \text{drug})$.
    Concrete paths (i.e., comprising specific entity sequences) matching the meta paths are sampled from the knowledge graph accordingly, e.g., $(\text{female reproductive system} \to \text{expression present} \to \text{SLC12A6} \to \text{target} \to \text{potassium chloride})$.

    \item \textbf{Generate memorization tasks.} We decompose a sampled path into fact triplets, the minimal units of knowledge in our setting and the fine-tuning material.
    For each fact triplet, an LLM (DeepSeek, \texttt{deepseek-chat}) or a predefined template phrases the memorization question.
    For intersection tasks, noise facts of both head entities are added to the list.

    \item \textbf{Generate generalization tasks.} From the same path, the generalization question is phrased in the same way.
\end{enumerate}

\noindent The LLM-based method yields diverse and more natural QA pairs, while the template-based method ensures consistency and controllability for downstream interpretability studies.

\subsection{Knowledge Novelty}
\label{sec:appendix_data_stats}

\begin{wraptable}{r}{0.52\linewidth}
  \vspace{-1.2em}
  \centering
  \setlength{\tabcolsep}{4pt}
  \caption{\textbf{Knowledge novelty.} Accuracy (\%) of the base models on 1{,}000 memorization questions, before fine-tuning.}
  \label{tab:prior_knowledge}
  \begin{tabular}{@{}lcc@{}}
    \toprule
    Model & \textsc{STaRK-Prime} & \textsc{STaRK-MAG} \\
    \midrule
    Qwen2.5-1.5B & 4.2 & 5.5 \\
    Qwen2.5-3B   & 3.8 & 6.0 \\
    Qwen2.5-7B   & 4.4 & 5.8 \\
    \midrule
    LLaMA-3.2-1B & 4.9 & 5.4 \\
    LLaMA-3.2-3B & 3.8 & 5.0 \\
    LLaMA-3.1-8B & 4.8 & 5.5 \\
    \bottomrule
  \end{tabular}
  \vspace{-0.8em}
\end{wraptable}

To check that the injected facts are new to the models, we ask each base model 1{,}000 randomly sampled memorization questions of each dataset before any fine-tuning.
Table~\ref{tab:prior_knowledge} reports the accuracy: it stays between 3.8\% and 4.9\% on \textsc{STaRK-Prime} and between 5.0\% and 6.0\% on \textsc{STaRK-MAG}, and it does not grow with model size, as the 7B and 8B models score about as low as the smallest model of their family.
The base models therefore do not know the injected facts in advance, and what they answer after fine-tuning comes from the injected facts.
Per-fact runs go one step further and discard every fact whose memorization question the base model already answers (Appendix~\ref{sec:appendix_experiment_setup}).

\section{Training and Evaluation Protocols}
\label{sec:appendix_experiment_setup}

All experiments inject facts with one of two protocols (Table~\ref{tab:protocols}).
\emph{Per-fact} runs inject only the supporting facts of a single question, so that the epoch at which this question becomes answerable is well defined; they measure the temporal lag and the permeation dynamics.
\emph{Multi-fact} runs inject a sample of 1{,}000 facts at once, as in practical knowledge injection, and are analyzed at convergence.

\begin{table}[h]
  \centering
  \small
  \renewcommand{\arraystretch}{1.12}
  \caption{\textbf{The two injection protocols.}}
  \label{tab:protocols}
  \begin{tabularx}{\linewidth}{@{}>{\raggedright\arraybackslash}p{2.3cm}>{\raggedright\arraybackslash}X>{\raggedright\arraybackslash}X@{}}
    \toprule
    & \textbf{Per-fact runs} & \textbf{Multi-fact runs} \\
    \midrule
    Used in & Fig.~\ref{fig:knowing_using_gap}, Fig.~\ref{fig:chaining_dynamics}, Table~\ref{tab:mcnemar_position}; App.~\ref{sec:appendix_scaling}, \ref{sec:appendix_permeation}
            & Tables~\ref{tab:patching_results} and~\ref{tab:prompting_patch_controls}, Fig.~\ref{fig:patching_position_distribution}, \S\ref{sec:lrsd}; App.~\ref{sec:appendix_scaling}, \ref{sec:appendix_mag}, \ref{sec:appendix_misalignment} \\
    Facts per run & the supporting facts of one question (noise facts included for intersection) & 1{,}000 (500 chaining or about 180 intersection questions) \\
    Questions & 100 per task, drawn with a fixed seed & a fixed random sample of facts per task \\
    Models & Qwen2.5-1.5B/3B, LLaMA-3.2-1B/3B; Qwen2.5-7B and LLaMA-3.1-8B (Fig.~\ref{fig:chaining_dynamics}, Table~\ref{tab:mcnemar_position}) & Qwen2.5-1.5B/3B/7B, LLaMA-3.2-1B/3B, LLaMA-3.1-8B \\
    Fine-tuning & LoRA or full fine-tuning & LoRA \\
    Epochs / batch / lr & 30 / 1 / $2{\times}10^{-5}$ & 50 / 10 / $10^{-4}$ \\
    Evaluation & after every epoch; one gold answer, correct if the output contains it or is contained in it & at the final epoch; exact match against any gold answer of the question \\
    \bottomrule
  \end{tabularx}
\end{table}

\noindent\textbf{Shared settings.}
Both protocols use AdamW with a constant learning rate and weight decay $0.01$, bf16 mixed precision, and a cross-entropy loss on the answer tokens only.
LoRA has rank 8, $\alpha=32$ and dropout $0.05$ on the $q,k,v,o,\mathrm{gate},\mathrm{up},\mathrm{down}$ projections of every layer.
Questions are answered by greedy decoding with at most 64 new tokens.
The STaRK-MAG runs follow the multi-fact protocol.

\noindent\textbf{Temporal lag.}
A per-fact run is discarded if the base model already answers one of its memorization questions before fine-tuning.
$T_{\text{mem}}$ is the first epoch at which the injected facts are memorized.
$T_{\text{gen}}$ is the first epoch from which the generalization question is answered correctly at 10 consecutive epochs.
Both are averaged over the runs that reach them; $A_{\text{mem}}$ and $A_{\text{gen}}$ are averaged over all runs at epoch 30.

\noindent\textbf{Self-patching.}
A state is the residual-stream output of a decoder block, read and overwritten on all tokens of the head entity of the first supporting fact.
In the converged maps of multi-fact runs (Tables~\ref{tab:patching_results}--\ref{tab:prompting_patch_controls}, Fig.~\ref{fig:patching_position_distribution}, App.~\ref{sec:appendix_misalignment}), a cell scores the greedy continuation by exact match; the oracle (``pat.'') counts a question as recovered if any layer pair, the diagonal included, answers it correctly.
In the permeation maps of per-fact runs (Fig.~\ref{fig:chaining_dynamics}, App.~\ref{sec:appendix_permeation}), a cell is the reciprocal rank of the answer under teacher forcing (from 0 to 1); the maps are computed at every epoch from one epoch before $T_{\text{mem}}$ to $T_{\text{gen}}$ for runs that generalize, and at every second epoch up to epoch 30 for runs that do not.

\noindent\textbf{Controls.}
The token-position ablation (Table~\ref{tab:mcnemar_position}) uses about 500 per-fact Qwen2.5-7B chaining runs that memorize their facts but never generalize, at epoch 30; the patched tokens are the head entity, \texttt{<BOS>}, \texttt{<EOS>}, the first or second relation, or a random position, and each position is compared with the random one by McNemar's test.
CoT prompting (Table~\ref{tab:prompting_patch_controls}) asks for two reasoning steps and reads the answer after ``Answer:''; outputs without this label count as wrong, and intersection is evaluated on 181 questions.
Irrelevant patching (Tables~\ref{tab:prompting_patch_controls} and~\ref{tab:control_self_patch_llama3b}) writes the head-entity state of another, randomly drawn question into the same layer, and reports the unpatched accuracy plus the mean over questions of the best gain over layers.

\noindent\textbf{Compute.}
Experiments ran on a server with eight NVIDIA A800 GPUs (80\,GB each), an Intel 128-core CPU and 512\,GB of memory.
A per-fact run takes 3--11 minutes on one GPU and a layer-wise self-distillation run 25--65 minutes; a multi-fact run with its layer-pair scans takes from about 5 hours to a day.
\section{Additional Results on the Knowing--Using Gap}
\label{sec:appendix_scaling}

This appendix supports Section~\ref{sec:lag_across_tasks}.
Table~\ref{tab:knowing_using_gap} lists the values plotted in Figure~\ref{fig:knowing_using_gap}.

\begin{table}[t]
  \centering
  \small
  \setlength{\tabcolsep}{3pt}
  \renewcommand{\arraystretch}{1.12}
  \begin{tabular*}{\linewidth}{@{\extracolsep{\fill}} l ccc ccc ccc ccc @{}}
    \toprule
      & \multicolumn{6}{c}{\textbf{Chaining}} & \multicolumn{6}{c}{\textbf{Intersection}} \\
    \cmidrule(lr){2-7} \cmidrule(lr){8-13}
      & \multicolumn{3}{c}{\textit{Lag (epochs)}} & \multicolumn{3}{c}{\textit{Accuracy}}
      & \multicolumn{3}{c}{\textit{Lag (epochs)}} & \multicolumn{3}{c}{\textit{Accuracy}} \\
    \cmidrule(lr){2-4} \cmidrule(lr){5-7} \cmidrule(lr){8-10} \cmidrule(lr){11-13}
    \textbf{Model}
      & $T_{\text{mem}}$ & $T_{\text{gen}}$ & $\boldsymbol{\Delta T}$ & $A_{\text{mem}}$ & $A_{\text{gen}}$ & $\boldsymbol{\Delta A}$
      & $T_{\text{mem}}$ & $T_{\text{gen}}$ & $\boldsymbol{\Delta T}$ & $A_{\text{mem}}$ & $A_{\text{gen}}$ & $\boldsymbol{\Delta A}$ \\
    \midrule
    \multicolumn{13}{@{}l}{\textit{LoRA}} \\
    Qwen-2.5-1.5B & 10.9 & 16.9 & \textbf{6.0} & 1.00 & 0.33 & \textbf{0.67} & 6.3 & 6.8 & \textbf{0.5} & 1.00 & 0.89 & \textbf{0.11} \\
    Qwen-2.5-3B   & 10.9 & 16.9 & \textbf{6.0} & 1.00 & 0.32 & \textbf{0.68} & 6.0 & 7.7 & \textbf{1.7} & 0.98 & 0.85 & \textbf{0.13} \\
    \addlinespace[2pt]
    LLaMA-3.2-1B  & 9.8 & 15.9 & \textbf{6.1} & 1.00 & 0.32 & \textbf{0.68} & 5.6 & 8.5 & \textbf{2.9} & 1.00 & 0.94 & \textbf{0.06} \\
    LLaMA-3.2-3B  & 10.4 & 15.0 & \textbf{4.6} & 1.00 & 0.30 & \textbf{0.70} & 5.8 & 8.0 & \textbf{2.2} & 0.98 & 0.88 & \textbf{0.10} \\
    \midrule
    \multicolumn{13}{@{}l}{\textit{Full fine-tuning}} \\
    Qwen-2.5-1.5B & 2.1 & 4.4 & \textbf{2.3} & 1.00 & 0.59 & \textbf{0.41} & 1.9 & 8.1 & \textbf{6.2} & 1.00 & 0.89 & \textbf{0.11} \\
    Qwen-2.5-3B   & 2.5 & 4.6 & \textbf{2.1} & 1.00 & 0.61 & \textbf{0.39} & 2.2 & 14.0 & \textbf{11.8} & 1.00 & 0.91 & \textbf{0.09} \\
    \addlinespace[2pt]
    LLaMA-3.2-1B  & 2.6 & 7.1 & \textbf{4.5} & 1.00 & 0.30 & \textbf{0.70} & 2.1 & 11.3 & \textbf{9.2} & 1.00 & 0.90 & \textbf{0.10} \\
    LLaMA-3.2-3B  & 2.4 & 7.9 & \textbf{5.5} & 1.00 & 0.32 & \textbf{0.68} & 2.3 & 12.4 & \textbf{10.1} & 1.00 & 0.92 & \textbf{0.08} \\
    \bottomrule
  \end{tabular*}
  \caption{\textbf{The Knowing--Using Gap across models and fine-tuning schemes (STaRK-Prime).}
  Each fact is injected in its own 30-epoch fine-tuning run; entries average over runs.
  $T_{\text{mem}}$: first epoch at which all injected facts are recalled; $T_{\text{gen}}$: first epoch after which the generalization query stays correct for 10 consecutive epochs (over runs that reach it); $A_{\text{mem}}$, $A_{\text{gen}}$: accuracy at the final epoch (for intersection, $A_{\text{mem}}$ is relative to the achievable maximum, since some memorization queries have several correct answers).
  The gap measures $\Delta T = T_{\text{gen}} - T_{\text{mem}}$ and $\Delta A = A_{\text{mem}} - A_{\text{gen}}$ are in bold.}
  \label{tab:knowing_using_gap}
\end{table}

\begin{figure}[h]
  \centering
  \begin{minipage}[c]{0.54\linewidth}
    \centering
    \includegraphics[width=\linewidth]{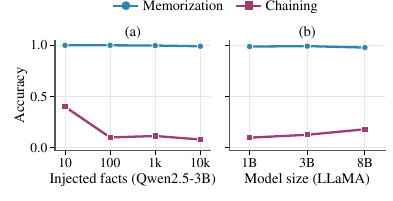}
  \end{minipage}\hfill
  \begin{minipage}[c]{0.42\linewidth}
    \caption{\textbf{Memorization and chaining accuracy across data and model size} (multi-fact protocol, final epoch).
    (a)~Qwen2.5-3B with 10 to 10{,}000 injected facts, i.e., 5 to 5{,}000 chaining questions.
    (b)~LLaMA-3.2-1B, LLaMA-3.2-3B, and LLaMA-3.1-8B with 1{,}000 facts (500 chaining questions).}
    \label{fig:scaling}
  \end{minipage}
\end{figure}

\noindent\textbf{Data size and model size.}
Figure~\ref{fig:scaling} repeats the multi-fact protocol while varying the number of injected facts (Qwen2.5-3B, 10 to 10{,}000 facts) and the model size (LLaMA-3.2-1B, LLaMA-3.2-3B, LLaMA-3.1-8B, 1{,}000 facts).
Memorization stays between 0.98 and 1.00 throughout, whereas chaining accuracy stays between 0.08 and 0.18 whenever at least 100 facts are injected; only the 10-fact run is higher, answering 2 of its 5 chaining questions.

\section{Additional Permeation Maps}
\label{sec:appendix_permeation}

\begin{figure}[p]
  \centering
  \includegraphics[width=\linewidth]{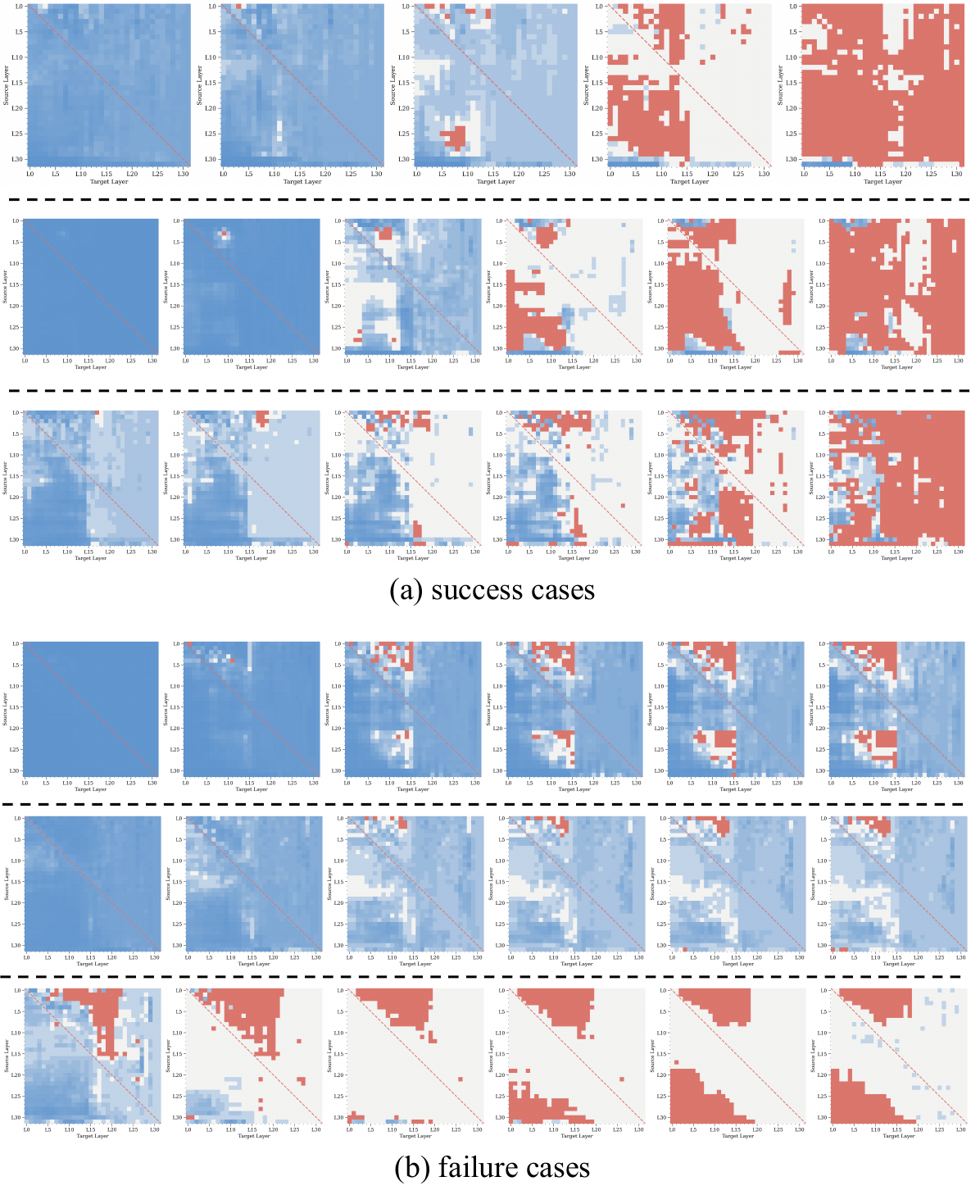}
  \caption{\textbf{Permeation maps of more per-fact chaining runs on LLaMA-3.1-8B.}
  Each row of maps follows one run across epochs, from left to right; in each map, rows are the source layer $l_{\text{src}}$ and columns the target layer $l_{\text{tgt}}$, and a cell is the reciprocal rank of the answer after patching, from 0 (blue) to 1 (red).
  (a)~Runs that generalize, from one epoch before $T_{\text{mem}}$ to $T_{\text{gen}}$; (b)~runs that do not.}
  \label{fig:more_patch_case_llama}
\end{figure}

\begin{figure}[p]
  \centering
  \includegraphics[width=\linewidth]{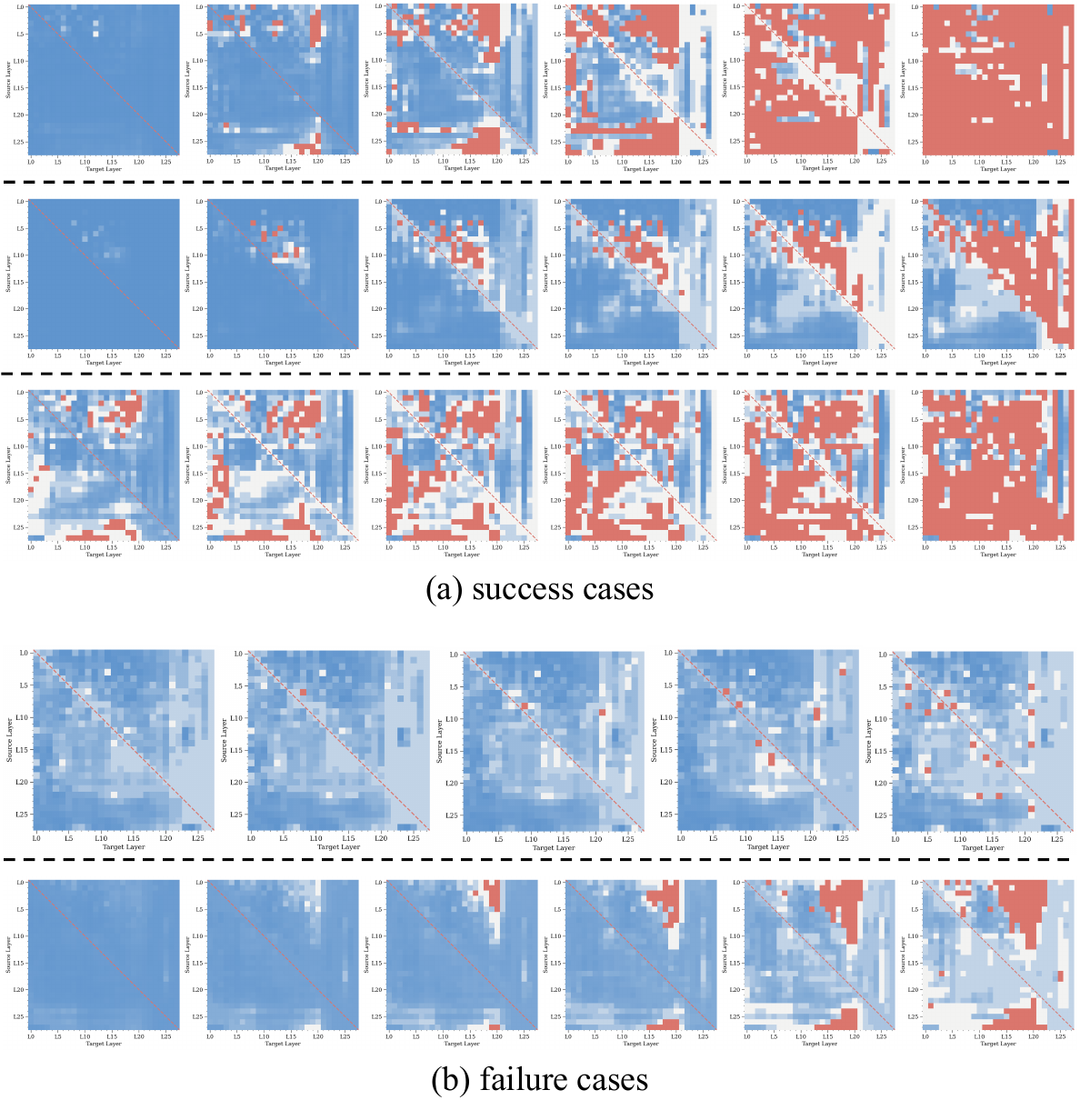}
  \caption{\textbf{Permeation maps of more per-fact chaining runs on Qwen2.5-7B}, as in Figure~\ref{fig:more_patch_case_llama}.
  (a)~Runs that generalize; (b)~runs that do not.}
  \label{fig:more_patch_case_qwen}
\end{figure}

This appendix supports Section~\ref{sec:permeation}.
Figures~\ref{fig:more_patch_case_llama} and~\ref{fig:more_patch_case_qwen} extend Figure~\ref{fig:chaining_dynamics} to more per-fact chaining runs of LLaMA-3.1-8B and Qwen2.5-7B (Appendix~\ref{sec:appendix_experiment_setup}).
Each row follows one run across epochs; a cell is the reciprocal rank of the answer after patching layer $l_{\text{src}}$ into layer $l_{\text{tgt}}$, from 0 (blue) to 1 (red), and the diagonal is the unpatched model.
The pattern of Figure~\ref{fig:chaining_dynamics} recurs: in runs that generalize, off-diagonal regions that fully restore the answer appear around $T_{\text{mem}}$ and reach the diagonal by $T_{\text{gen}}$; in runs that do not, such regions appear but the diagonal is never restored.
Counted over all scanned runs that generalize, some off-diagonal layer pair already fully restores the answer at $T_{\text{mem}}$ in 73\% of the LLaMA runs (26 runs) and 63\% of the Qwen runs (43 runs), whereas the unpatched model answers in 12\% and 5\% of them.

\section{Cross-Domain Replication on STaRK-MAG}
\label{sec:appendix_mag}

\begin{figure}[h]
  \centering
  \includegraphics[width=\linewidth]{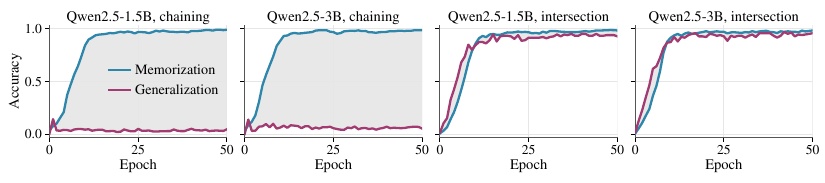}
  \caption{\textbf{Training curves on STaRK-MAG} (Qwen2.5, multi-fact protocol, accuracy after every epoch).
  The shaded area is the difference between memorization and generalization accuracy.}
  \label{fig:mag_training_dynamics}
\end{figure}

This appendix supports the STaRK-MAG columns of Table~\ref{tab:patching_results} (Section~\ref{sec:manual_alignment}).
STaRK-MAG facts are injected with the multi-fact protocol (Appendix~\ref{sec:appendix_experiment_setup}), and the patching evaluation uses 500 chaining and 166 intersection questions.
Figure~\ref{fig:mag_training_dynamics} shows the training curves of the two Qwen2.5 models.
Chaining reproduces the gap found on STaRK-Prime: memorization exceeds 0.95 by epoch 12--16, whereas chaining accuracy never exceeds 0.10 after epoch 5 and ends at 0.05.
Intersection accuracy rises together with memorization and ends at 0.93--0.96, again as on STaRK-Prime.

\section{Additional Controls for Self-Patching}
\label{sec:appendix_misalignment}

This appendix supports the controls of Section~\ref{sec:controls_prompting}.

\begin{wraptable}{r}{0.4\linewidth}
  \vspace{-1.1em}
  \centering
  \small
  \setlength{\tabcolsep}{4pt}
  \caption{\textbf{Irrelevant patching on LLaMA-3.2-3B}, as in Table~\ref{tab:prompting_patch_controls}.}
  \label{tab:control_self_patch_llama3b}
  \begin{tabular}{@{}lcc@{}}
    \toprule
    & Chaining & Intersection \\
    \midrule
    No patching         & 0.126 & 0.815 \\
    Irrelevant patching & 0.207 & 0.915 \\
    Self-patching       & \textbf{0.404} & \textbf{0.969} \\
    \bottomrule
  \end{tabular}
  \vspace{-0.8em}
\end{wraptable}

\noindent\textbf{Irrelevant patching on LLaMA.}
Table~\ref{tab:control_self_patch_llama3b} repeats the irrelevant-patching control of Table~\ref{tab:prompting_patch_controls} on LLaMA-3.2-3B.
As on Qwen2.5, writing in the state of an unrelated fact helps far less than writing the fact's own state (0.207 against 0.404 on chaining).

\begin{figure}[h]
  \centering
  \includegraphics[width=0.83\linewidth]{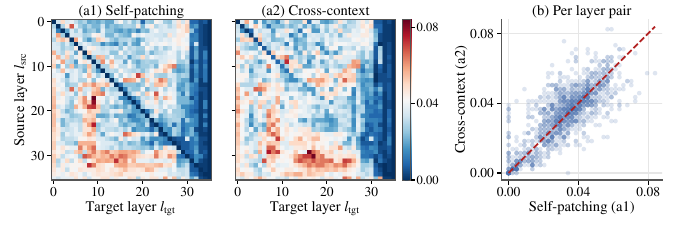}
  \caption{\textbf{Cross-context consistency of self-patching} (Qwen2.5-3B, 1{,}000 \textsc{STaRK-Prime} chaining facts, epoch 50).
  A cell $(l_{\text{src}},l_{\text{tgt}})$ is the exact-match rate on the chaining questions the model does not answer without patching, after the head-entity state at layer $l_{\text{src}}$ of a source prompt is written into layer $l_{\text{tgt}}$ of the chaining prompt $P_{\text{gen}}$; both maps share one colour scale.
  \textbf{(a1)}~Self-patching: the source is $P_{\text{gen}}$ itself, so the diagonal is the unpatched model.
  \textbf{(a2)}~Cross-context patching: the source is the memorization query $P_{\text{mem}}$ of the first-hop fact.
  \textbf{(b)}~One point per layer pair, value in (a1) against value in (a2); dashed: $y=x$.}
  \label{fig:mem_self_patching_comparison}
\end{figure}

\noindent\textbf{Cross-context patching.}
The prompting and perturbation controls in \S\ref{sec:controls_prompting} leave one alternative open: the gains of self-patching might come from artifacts of the chaining prompt rather than from the injected fact.
We therefore take the source state from a different context, the single-hop memorization query $P_{\text{mem}}$ of the first-hop fact, which is a training prompt and contains the same head entity, and write its head-entity state into the chaining prompt $P_{\text{gen}}$ with the same scan over $(l_{\text{src}},l_{\text{tgt}})$.
Figure~\ref{fig:mem_self_patching_comparison} compares this cross-context map with the self-patching map on Qwen2.5-3B at convergence.
The two maps share the same layout: gains come mostly from late source layers written into earlier target layers, and target layers near the output (the blue columns at the right) never help.
Their per-pair values also lie along $y=x$ (panel b; Pearson $r=0.80$ over the off-diagonal pairs), so the effect of a layer pair barely depends on which prompt supplies the state.
The useful representation is thus already formed on the memorization prompt, which supports that self-patching relocates injected knowledge rather than noise.

\section{Layer-wise Self-Distillation Details}
\label{sec:appendix_lrsd}

This appendix supports Section~\ref{sec:lrsd}; LRSD uses the multi-fact protocol of Appendix~\ref{sec:appendix_experiment_setup}.

\noindent\textbf{Data.}
LRSD is trained on the 1{,}000-fact \textsc{STaRK-Prime} samples of the oracle experiments in Section~\ref{sec:manual_alignment}, one sample per model family.
Chaining uses 500 two-hop instances, whose 409 (Qwen) and 427 (LLaMA) distinct questions form the evaluation set.
Every training example is a single-hop memorization query; no generalization question is ever trained on.

\noindent\textbf{Training.}
LoRA with rank 8, $\alpha=32$ and dropout $0.05$ on $\{q,k,v,o,\mathrm{gate},\mathrm{up},\mathrm{down}\}_{\mathrm{proj}}$ of every block; fp32 weights with bf16 autocast; AdamW with constant learning rate $10^{-4}$, weight decay $0.01$ and gradient-norm clipping at $1.0$; batch size 10; 50 epochs; cross-entropy on the answer tokens only.
The seeds $\{848,1,2\}$ set the LoRA initialization, dropout and data order, while the fact sample is fixed.
$h^{l}$ is the output of decoder block $l$ (0-indexed), read with a forward hook, and $(l_{\text{src}},l_{\text{tgt}})=(0.75L,0.5L)$ gives $(27,18)$ for Qwen2.5-3B ($L=36$), $(21,14)$ for Qwen2.5-1.5B and LLaMA-3.2-3B ($L=28$), and $(12,8)$ for LLaMA-3.2-1B ($L=16$).
The entity span is found by matching the entity string in the tokenized prompt; the few prompts without a match ($\leq$1.3\%) get no LRSD term.
$\mathcal{L}_{\text{LRSD}}$ is averaged over the prompts of a batch and is active from the first step.
A run takes 25--65 minutes on one A800 GPU, with two runs sharing a GPU.

\noindent\textbf{Evaluation.}
Before training and after every epoch, we decode each distinct generalization question greedily (at most 64 new tokens) and count it as correct when the whole output matches one of its gold answers, ignoring case and surrounding whitespace; memorization uses the same rule on all memorization queries every 5 epochs.
No patching is applied at any point.
A run is summarized by its mean exact match over epochs 30--50.
Each LRSD run is compared with the plain fine-tuning run of the same seed, and the 12 seed-paired differences (four models $\times$ three seeds) are tested with a two-sided exact sign-flip test.

\begin{table}[h]
  \centering
  \small
  \setlength{\tabcolsep}{4.5pt}
  \renewcommand{\arraystretch}{1.1}
  \begin{tabular}{l c c c c c c c}
    \toprule
    & & \multicolumn{3}{c}{\textbf{Chaining, epochs 30--50}} & \multicolumn{3}{c}{\textbf{Memorization, epoch 50}} \\
    \cmidrule(lr){3-5}\cmidrule(lr){6-8}
    \textbf{Model} & $l_{\text{src}}\to l_{\text{tgt}}$ & $\lambda=0$ & $\lambda=0.1$ & $\lambda=1$ & $\lambda=0$ & $\lambda=0.1$ & $\lambda=1$ \\
    \midrule
    Qwen2.5-1.5B & 21$\to$14 & .085{\scriptsize$\pm$.008} & .116{\scriptsize$\pm$.005} & .166{\scriptsize$\pm$.002} & .985 & .992 & .987 \\
    Qwen2.5-3B   & 27$\to$18 & .096{\scriptsize$\pm$.011} & .138{\scriptsize$\pm$.005} & .187{\scriptsize$\pm$.013} & .995 & .991 & .978 \\
    LLaMA-3.2-1B & 12$\to$8  & .089{\scriptsize$\pm$.004} & .095{\scriptsize$\pm$.006} & .106{\scriptsize$\pm$.009} & .977 & .988 & .987 \\
    LLaMA-3.2-3B & 21$\to$14 & .132{\scriptsize$\pm$.002} & .148{\scriptsize$\pm$.005} & .157{\scriptsize$\pm$.007} & .994 & .986 & .982 \\
    \bottomrule
  \end{tabular}
  \caption{\textbf{LRSD results.}
  Exact match without patching, mean $\pm$ s.d.\ over seeds $\{848,1,2\}$.
  Seed-paired comparisons with $\lambda=0$, pooled over the four models: LRSD is better in 12 of 12 pairs at both $\lambda=0.1$ and $\lambda=1$ ($p<0.001$).}
  \label{tab:lrsd_full}
\end{table}

\noindent\textbf{Full results.}
Table~\ref{tab:lrsd_full} lists all settings.

\noindent\textbf{Choice of layer pair and position.}
Table~\ref{tab:lrsd_pilot} lists the pilot runs on Qwen2.5-3B chaining from which the configuration was fixed.
Among the head-entity pairs, 27$\to$18 ($0.75L\to0.5L$) scored best on seed 848 and was then transferred to the other models by relative depth.
Two controls with five seeds each show that the gain needs both ingredients: the same loss on the tokens just before the entity gives no gain, and aligning two late layers (32$\to$25) gives less than a third of the gain of 18$\to$11.
Pulling layer 18 toward layer 9 (9$\to$18) destroys memorization, also at $\lambda=0.1$ (0.51 at epoch 50).

\begin{table}[h]
  \centering
  \small
  \setlength{\tabcolsep}{5pt}
  \renewcommand{\arraystretch}{1.1}
  \begin{tabular}{l l c c c}
    \toprule
    \textbf{Pair} $l_{\text{src}}\to l_{\text{tgt}}$ & \textbf{Positions} & \textbf{Seeds} & \textbf{Gen.\ (epochs 30--50)} & \textbf{Mem.\ (epoch 50)} \\
    \midrule
    --- (plain fine-tuning) & --- & 5 & .102 & .995 \\
    \midrule
    27$\to$18 & head entity & 3 & .187 & .978 \\
    27$\to$9  & head entity & 1 & .136 & .962 \\
    18$\to$11 & head entity & 5 & .153 & .954 \\
    11$\to$18 & head entity & 4 & .159 & .943 \\
    9$\to$18  & head entity & 1 & .017 & .261 \\
    \midrule
    32$\to$25 & head entity & 5 & .117 & .988 \\
    18$\to$11 & tokens just before the entity & 5 & .101 & .994 \\
    \bottomrule
  \end{tabular}
  \caption{\textbf{Pilot runs for the layer pair and position} (Qwen2.5-3B, chaining, $\lambda=1$).
  $l_{\text{src}}\to l_{\text{tgt}}$: the layer-$l_{\text{tgt}}$ states are pulled toward the stop-gradient layer-$l_{\text{src}}$ states.
  Exact match without patching, mean over seeds.
  27$\to$18 was chosen on seed 848 (0.201, against at most 0.163 for the other pairs); its other two seeds are those of Table~\ref{tab:lrsd_full}.}
  \label{tab:lrsd_pilot}
\end{table}

\noindent\textbf{Alignment analysis.}
The alignment carries over to the two-hop chaining questions, on which LRSD is never trained.
There, the relative distance $\lVert h^{l_{\text{tgt}}}-h^{l_{\text{src}}}\rVert_2^2/\lVert h^{l_{\text{src}}}\rVert_2^2$ at the head entity falls from 0.34--0.71 before training to about 0.001 at epoch 50 under $\lambda=1$ (0.009--0.041 under $\lambda=0.1$), whereas plain fine-tuning slightly increases it (e.g., from 0.71 to 0.78 on LLaMA-3.2-1B).

\section{Limitations}
\label{sec:appendix_responsible}

Self-patching is deliberately constrained: we intervene at a fixed anchor position and move representations across layers. Injected knowledge may also distribute across multiple positions or be redundantly encoded, so our estimates may understate the full recoverable headroom.

Our oracle best-pair results should be interpreted as a diagnostic upper bound. Layer-wise self-distillation (Section~\ref{sec:lrsd}) indeed closes part of this headroom by demonstrating real improvements in generalization, but it does not yet fully close the gap.

LRSD, while can be put into practical use, is mainly a proof-of-concept method and still with naive implementation: It's not the focus of this work, we hope to prove that even with a simple selection of pairing and internal alignment will improve generalization of knowledge.

Finally, our mechanistic analysis operates at the token-layer level. Finer-grained localization to specific attention heads or MLP sublayers could further refine the knowledge--circuit misalignment hypothesis and inform more targeted interventions.

\end{document}